\documentclass[trsc,nonblindrev]{informs3} 
\usepackage{enumitem}

\OneAndAHalfSpacedXI 

\usepackage{natbib}
 \bibpunct[, ]{(}{)}{,}{a}{}{,}%
 \def\bibfont{\small}%
\usepackage{xcolor}
\usepackage{bm}
\usepackage{amsmath}
\usepackage{booktabs}
\usepackage{tabularx}
\usepackage{threeparttable}
\usepackage{multicol}
\usepackage{multirow}
\usepackage[ruled,vlined]{algorithm2e}
\usepackage{graphicx}
\usepackage{subcaption}
\allowdisplaybreaks[4]
\usepackage[colorlinks=true, allcolors=blue]{hyperref}
\usepackage{amssymb}

\makeatletter
\def\@currsize{\selectfont}
\makeatother

\DontPrintSemicolon
\SetAlCapFnt{\small}
\SetAlCapNameFnt{\small}
\SetKwInput{KwIn}{Input}
\SetKwInput{KwOut}{Output}

\TheoremsNumberedThrough     

\EquationsNumberedThrough    

\newcommand{\R}{\mathbb{R}}
\begin{document}





\TITLE{A Retrieval-Enhanced Transformer for Multi-Step Port-of-Call Sequence Prediction in Global Liner Shipping}

\ARTICLEAUTHORS{%
\AUTHOR{Yanzhao Su, Fang He\thanks{Corresponding author. Email: \EMAIL{fanghe@tsinghua.edu.cn}}}
\AFF{Department of Industrial Engineering, Tsinghua University, Beijing 100084, P.R. China,}
\AUTHOR{Yineng Wang}
\AFF{Department of Logistics and Maritime Studies, The Hong Kong Polytechnic University, Hong Kong,}
} 

\ABSTRACT{%
Accurate multi-step port-of-call sequence prediction is vital for enhancing tactical resource orchestration and logistical efficiency, yet existing methods often struggle with the limited accessibility and unreliability of voyage schedules, alongside the inability of AIS data to provide visibility beyond the immediate next port. To address these challenges, this study proposes a generalized Connectivity-Constrained and Retrieval-Enhanced (CCRE) deep learning framework for multi-step port sequence prediction. Inspired by the paradigm of Retrieval-Augmented Generation in modern deep learning, the framework introduces a retrieval-enhanced historical encoder that explicitly queries a global maritime database to aggregate contextually similar navigational precedents. By transforming these scenarios into a candidate-level semantic representations, the model effectively compensates for data sparsity in long-tail routes and resolves routing ambiguities. Integrating this with a Transformer-based trajectory encoder, the architecture executes adaptive ``middle fusion" via cross-attention to dynamically shift the predictive reliance from real-time kinematics for short-term accuracy to historical context for long-term strategic stability. To ensure superior sequence-level coherence, we formulate forecasting as a joint sequence generation problem using an autoregressive Transformer decoder enriched with Scheduled Sampling and Gumbel-Softmax relaxation. This joint optimization facilitates end-to-end gradient propagation to mitigate error accumulation while topology masks strictly enforce maritime network reachability to eliminate operationally infeasible routes. Evaluated on a global dataset, CCRE achieves a 72.3\% first-destination accuracy and a 61.4\% average three-step accuracy, outperforming established baselines—specifically CatBoost and Long Short-Term Memory—by average margins of 12.6\% and 11.3\%, respectively. Feature ablation and qualitative case studies further corroborate the model's scalability and its ability to capture complex routing patterns across diverse international trade lanes.
}%


\KEYWORDS{Multi-step prediction, AIS data, Retrieval-Enhanced generation, Connectivity constraints, Liner shipping network}

\maketitle

%

\section{Introduction}

In the global liner shipping industry, the ability to foresee a vessel’s future port-of-call sequence is a cornerstone of operational efficiency, as a longer prediction horizon provides significant advantages to various maritime practitioners. For carriers, the complexities of modern alliance structures and vessel-sharing agreements mean that cargo movement is rarely a solitary endeavor. Because a single shipment often involves swapping or purchasing slots from other operators, a carrier’s logistical success depends on the reliability of their partners' schedules. Having a predictive, multi-step view of various carriers' port sequences allows for more robust cargo planning and transshipment synchronization. It transforms a reactive "wait-and-see" approach into a proactive strategy, enabling carriers to optimize route selection, minimize dwell times at transshipment hubs, and maintain service integrity even when individual vessels within a network face delay. For port and terminal operators, the value of a long-term horizon lies in tactical resource synchronization. While day-to-day berth allocation is highly dynamic, the broader strategic orchestration—such as capacity scaling, yard space optimization, heavy equipment maintenance and seasonal labor supply adjustments—must be planned 1 to 2 months in advance. This period often covers a vessel’s rotation through multiple ports within a region. Relying solely on "next-port" data creates a "reactive bottleneck," where terminals are blind to the cumulative delays building up upstream. 

While the planned voyage schedules disseminated by carriers under standards \citet{dcsa2024ovs} offer a straightforward source of sequence information, these records are often difficult to access and frequently inaccurate due to manual entry errors, omissions, or delayed updates \citep{IALA2016, yang2019big}. Furthermore, operational disruptions such as port skipping and rerouting introduce systematic deviations between planned schedules and realized port calls \citep{zhang2023liner} rendering static schedules a "nominal" rather than "predictable" truth. Our empirical analysis of Maersk’s 2024 operational data further underscores this discrepancy. Despite Maersk leading the industry with a 57.9\% schedule reliability rating in late 2024, a granular comparison between its official schedules and 38,785 realized port calls reveals a significant "predictability gap." Only 69.95\% of actual arrivals could be successfully matched to the planned schedule positions. Within these matched samples, while the accuracy for the immediate "next-port" was approximately 54.56\%, the accuracy for a multi-step horizon—specifically, three consecutive future calls—plummeted to a mere 17.70\%. 

Given the limited accessibility and inherent unreliability of schedule data, Automatic Identification System (AIS) data has become a primary alternative for vessel trajectory prediction. Originally mandated for navigational safety, AIS now serves as a ubiquitous sensor network for high-frequency vessel status updates \citep{harati2007automatic}, and its standardized implementation has led to wide adoption across maritime research fields \citep{Tu2018exploring}. However, a critical limitation remains: AIS signals are inherently kinematic rather than intentional. Although AIS provides snapshots of a vessel’s current kinematic state and its next-port intention, it fails to capture the long-horizon itinerary or the latent dependencies between multiple future calls.

As this specific problem remains insufficiently explored, we summarize the remaining challenges into three critical dimensions. First, multi-step port-of-call prediction faces significant hurdles in capturing complex sequential dependencies and maintaining long-term routing logic consistency across global shipping networks. Since vessel behaviors often involve periodic shipping patterns or transitions between distinct service lines, robust sequential dependency modeling is essential to prevent the catastrophic error accumulation that typically occurs as the prediction horizon extends. Second, this complexity is further compounded by the inherent data sparsity and imbalanced long-tail distribution of maritime traffic, where a minority of popular routes dominate the volume while a vast number of rare routes lack sufficient data for reliable pattern extraction. Such imbalance necessitates innovative mechanisms to aggregate global navigational precedents to resolve routing ambiguities on these infrequent corridors. Third, effectively synthesizing disparate information sources—specifically high-resolution vessel kinematics and global navigational precedents—remains a formidable architectural challenge. A global-wise robust predictive framework requires the capacity to execute dynamic feature-level fusion, allowing for a strategic shift in focus as the prediction horizon extends. This necessitates a sophisticated encoding mechanism that can distill real-time trajectory dynamics while adaptively integrating them with retrieved historical context to ensure that short-term forecasts remain grounded in immediate vessel states, while long-term sequences are stabilized by established routing logic to maintain physical feasibility.

To address these gaps in multi-step port-of-call prediction, we propose a generalized framework encompassing a Connectivity-Constrained and Retrieval-Enhanced (CCRE) deep learning network for multi-step port sequence prediction based only on highly accessible AIS data. In contrast with the prior literature, our study bridges the gap between short-horizon port prediction and long-horizon multi-step voyage visibility. Our main contributions are summarized as follows:
\begin{itemize}
    \item Unlike single-step classification frameworks that neglect sequential dependencies, we employ an autoregressive Transformer decoder enriched with Scheduled Sampling and Gumbel-Softmax relaxation to conduct multi-step port-of-call prediction. This synergy facilitates end-to-end gradient propagation across the entire prediction horizon, effectively mitigating the inherent challenge of error accumulation and ensuring long-term routing logic consistency in global liner shipping networks.
    

    \item Inspired by the paradigm of Retrieval-Augmented Generation (RAG) in modern deep learning, the framework introduces a Retrieval-Enhanced historical encoder that explicitly queries a global maritime database to aggregate contextually similar navigational precedents. Unlike previous clustering-based models relying on isolated historical data, this module utilizes a dual-metric similarity mechanism, combining Jaccard Similarity to capture set-level global routing intent with the Positional Match Rate (PMR) for precise sequential alignment. By transforming these scenarios into a candidate-level semantic representations, the model effectively compensates for data sparsity in long-tail routes and resolves routing ambiguities that instantaneous kinematic states alone cannot disambiguate.

    \item Furthermore, the architecture employs a Transformer-based trajectory encoder to distill high-resolution AIS kinematics and operational contexts into a latent representation. Departing from traditional late-stage weighting or stacking methods, the proposed modeling framework executes adaptive ``middle fusion" at the feature level via a cross-attention module. This enables the model to dynamically shift its focus across the prediction horizon—relying on real-time kinematics for short-term accuracy while transitioning to historical context for long-term strategic stability—thereby ensuring robust and physically feasible multi-step sequence generation.


    \item The predictive effectiveness and generalization capability of the proposed framework are validated on a comprehensive global-scale liner shipping network rather than limited regional corridors. Extensive empirical experiments demonstrate that the model achieves an accuracy of 72.3\% for the first destination prediction, surpassing established baselines—specifically advanced tree-based ensembles such as CatBoost and deep sequence models like Long Short-Term Memory—by an average margin of 12.6\%. Furthermore, the framework attains an average accuracy of 61.4\% across a three-step prediction horizon, outperforming comparative benchmarks by an average of 11.3\%. Feature ablation confirms that the model adaptively shifts reliance from real-time kinematic features for short-term predictions to historical context for long-term forecasts, while qualitative case studies highlight its capability to capture both highly regular cyclic patterns and complex, partially regular trade lanes. These results corroborate the model's scalability and its superior ability to generalize across diverse international trade lanes compared to established baselines

\end{itemize}

The remainder of this paper is organized as follows. Section~\ref{sec:literature} introduces and reviews the state-of-the-art literature. Section~\ref{sec:methods} formalizes the research problem, outlines methodological approaches including preprocessing techniques, the model architecture, and evaluation metrics. Section~\ref{sec:design} describes the dataset details and experimental setup. Section~\ref{sec:experiment} presents a comparison of results between the proposed model and baseline models, along with ablation experiments and case study results for our model. Finally, Section~\ref{sec:conclusion} concludes the study and outlines directions for future work.

\section{Literature Review}\label{sec:literature}
This section first provides a systematic review of existing methodologies in prediction of maritime transportation, including single-step destination prediction, multi-step trajectory forecasting, and estimated time of arrival (ETA) prediction. Then, we identify the existing research gaps regarding multi-step port-call sequence prediction, which motivate the proposed framework.

\subsection{Forecasting problems in maritime transportation}


\subsubsection{Single-step destination prediction.}

A substantial body of work formulates vessel behavior prediction as the classification of a single future port, typically the next immediate call or the final destination, given partial trajectories and static vessel attributes. This problem is predominantly addressed using classical machine learning models and trajectory similarity metrics.

In the domain of feature-based classification, researchers treat prediction instances as independent events. \citet{Karatas2020TrajectoryAIS} compared multiple supervised models such as Random Forest (RF) and Support Vector Regression to address arrival port classification and arrival time regression jointly. Similarly, on specific trade corridors, \citet{lloret2025cross} employed artificial neural networks with kinematic and contextual features derived from the Automatic Identification System to predict the next destination and estimated time of arrival for cross-Pacific container ships. At the global scale, \citet{Kim2023WAY} reformulated worldwide trajectories as nested spatial-grid sequences and proposed the WAY model, which utilizes channel-aggregative self-attention to predict the destination port days to weeks in advance. Despite the long temporal look-ahead, the output in these studies remains a single port label rather than a sequence of future calls.

Complementary to classification, similarity-based approaches leverage historical trajectory matching to extrapolate future destinations. Fundamental work in this area involves quantifying similarity via metrics like Dynamic Time Warping (DTW) or shape-based clustering. \citet{zhang2020randomforest} advanced this by building a global database and learning similarity features via RF, significantly outperforming benchmark metrics. Hybrid architectures have also been proposed. For instance, \citet{yin2022vessel} developed a stacking framework combining Bayesian neural networks with a dynamic clustering-based core track similarity module. To address scalability issues, recent works employ higher-level route abstractions. \citet{Pallotta2013TREAD} introduced a framework to learn typical routes for probabilistic classification, while \citet{Zygouras2024EnvClus} constructed mobility graphs to capture inter-port pathways. While effective for single targets, these methods often require long observation windows to establish confident matches.

\subsubsection{Multi-step trajectory and route forecasting.}

In parallel to destination inference, a substantial literature has emerged on multi-step prediction in trajectory space. Unlike port-level prediction, these approaches utilize deep sequence models to forecast future kinematic states such as positions, grid cells, or continuous paths.

The application of sequence to sequence (Seq2Seq) models to maritime data was pioneered by \citet{Nguyen2018Seq2SeqTraj}, who discretized maritime areas into spatial grids to predict future cell sequences. In these grid-based models, ports are merely derived endpoints inferred when the predicted path intersects port regions. To capture the temporal dependencies of continuous movement, Recurrent Neural Networks (RNN) have become the standard. \citet{Yang2022DLSTM} combined trajectory clustering with Long Short-Term Memory (LSTM) encoder-decoder models for ocean-wide track forecasting. Subsequent research has refined this approach to capture complex spatio-temporal structures. \citet{gao2021novel} integrated cubic-spline interpolation with historical reference points to stabilize predictions during maneuvering, while others have adopted bidirectional and convolutional architectures to enhance structural capture \citep{park2021BiLSTMSpectral,wu2023ConvLSTMSeq2Seq}.

Recent advancements incorporate attention mechanisms to handle long-horizon dependencies. \citet{Li2024ACoAttLSTM} introduced attention-augmented networks to refine feature focusing, and \citet{yin2025LSTMMultiScaleAttn} utilized multi-scale convolutions with attention mechanisms. Furthermore, physics-informed neural networks have been proposed to embed kinematic dynamics into predictors, ensuring physical plausibility \citep{Alam2025PINN}. While these methods demonstrate methodological maturity in trajectory space, they focus on micro-level movement tendencies rather than the macro-level topological transitions between ports.

\subsubsection{ETA prediction.}

The third major focus is prediction of the estimated time of arrival once the destination port is given or has been inferred. This task is often coupled with destination prediction using both neural and ensemble methods.

The line of work from the DEBS 2018 Grand Challenge combined ensemble classifiers for destination with neural networks for arrival time prediction from streaming data \citep{Bodunov2018DEBSChallenge}. More recent studies explore tree ensembles and hybrid frameworks using features derived from the Automatic Identification System. For instance, \citet{Arbabkhah2024HoustonETA} used XGBoost to predict arrival times in narrow waterways, while \citet{saber2025HybridETA} proposed a hybrid machine-learning approach achieving low error rates at seaports. Integration of these predictions into operational decision making has been studied in berth allocation and port logistics contexts, where forecasts are embedded into optimization models \citep{chu2026BerthAllocVAT}. Across these works, the future port of interest is a single known or predicted entity, and the sequence of subsequent ports is not modeled.

\subsection{Existing research gaps}

Despite the extensive literature on maritime destination and trajectory forecasting, a critical review reveals significant gaps when addressing the specific problem of predicting multi-step future port-call sequences. The limitations of current methodologies can be synthesized into three primary dimensions as follows.

First, single-step classifiers exhibit inherent myopia and limited global topological awareness. 
While classic machine learning models (e.g., RF, XGBoost) excel at single-step destination prediction or ETA regression \citep{Karatas2020TrajectoryAIS, lloret2025cross}, they are inherently designed for ``one-shot" tasks. The prediction instances are treated as independent events and the intricate connectivity constraints for multi-step prediction in the global maritime network are neglected. 

Second, there is a modality mismatch in multi-step continuous trajectory prediction and discrete future port sequence prediction. 
Though deep sequence models (e.g., LSTM, Seq2Seq, Transformers) have become the standard for forecasting continuous vessel trajectories \citep{Nguyen2018Seq2SeqTraj, Kim2023WAY}, the port sequence prediction requires specific network design to capture the critical global information and behavioral patterns as well as to output feasibly in line with the port networks.


Third, considering the similarity of port-of-call sequence on a global scale confronts scalability and generalization issues. The computational cost of pairwise similarity or searching massive historical archives become prohibitive for global real-time applications. Moreover, learning approaches relying on trajectory similarity and route matching \citep{Pallotta2013TREAD, Zygouras2024EnvClus, zhang2020randomforest} struggle with the long-tail distribution of maritime traffic and often fail to generalize to rare routes or sparse observation windows.

We propose the CCRE neural network to overcome these deficiencies. The model adopts an autoregressive Transformer decoder with Gumbel-Softmax to facilitate gradient propagation across multiple time steps, thereby stabilizing long-horizon predictions. To manage the immense search space of global shipping, it reformulates the retrieval process using connectivity constraints and semantic alignment of historical sequences. Finally, the integration of retrieval-enhanced learning serves to alleviate the impact of data sparsity and long-tail distributions.

\section{Methods}\label{sec:methods}
In this section, we propose a comprehensive methodological framework designed to address the discrepancy between raw maritime observations and long-horizon port-call forecasting. Firstly, Section~\ref{sec:formulation} formalizes the multi-step prediction task as a connectivity-constrained sequence generation problem, thereby establishing the global objective and the topological constraints that govern the output space. Then, Section~\ref{sec:data-processing} details the feature engineering pipeline, which transforms unstructured AIS signals into the structured semantic forms required by our problem definition. Finally, Section~\ref{sec:model} introduces the CCRE network composed of a trajectory encoder, a retrieval-enhanced historical encoder, an adaptive fusion module, and a connectivity-constrained decoder that executes the mapping from input features to feasible port sequences to solve the defined optimization task.

\subsection{Model Formulation}\label{sec:formulation}

To better illustrate our problem and articulate our optimization objectives, we first formalize the task of multi-step port call sequence prediction by establishing the structural foundation of the global liner shipping industry.

We model the maritime network as a directed graph $\mathcal{G}=(\mathcal{P}, \mathcal{E})$, where $\mathcal{P}=\{1, \dots, |\mathcal{P}|\}$ represents a finite vocabulary of global ports and an edge $(i, j) \in \mathcal{E}$ denotes a historically observed direct service between port $i$ and port $j$. This structure is mathematically encapsulated in an adjacency matrix $A \in \{0, 1\}^{|\mathcal{P}| \times |\mathcal{P}|}$, where $A_{ij} = 1$ if a direct connection exists. To accommodate variable-length sequences and ensure numerical stability, we augment the vocabulary with a sentinel token $\omega = |\mathcal{P}| + 1$, which serves as a placeholder for unknown or padding entries.

In order to bridge the gap between static networks and dynamic vessel behavior, we define a historical port context for each sample $s$ to capture the vessel's operational trajectory. This port-based operational history not only reduces the storage required for historical information but also partially reflects the vessel's navigational trends over a given past period:
\begin{equation}\label{eq:history_sequence}
    \bm{h_{s}} = (h_{s,1}, \dots, h_{s,K}) \in (\mathcal{P} \cup \{\omega\})^{K}, \quad o_{s} := h_{s,K},
\end{equation}
where $\bm{h_{s}}$ is a chronologically ordered sequence recording the $K$ most recent completed port calls. Each element $h_{s,k}$ represents a discrete port index or the sentinel token $\omega$, while the lookback window $K$ defines the depth of the observable operational memory. The terminal element $o_{s}$ is designated as the current origin port, representing the last validly observed berthing location before the prediction horizon. Beyond its role as a state descriptor, $o_{s}$ serves as a critical spatio-temporal anchor that provides dual-purpose initialization for both the similarity-based historical retrieval module and the reachability analysis in subsequent decoding stages.

Unlike generic sequence tasks, vessel movements are governed by a rigid shipping network topology. To prevent the generation of operationally infeasible routes, we define a step-dependent topological feasibility domain $\mathcal{R}_{h}(o_{s})$ for each prediction step $h$:
\begin{equation}\label{eq:reachability}
\begin{aligned}
    \mathcal{R}_h(o_s) &= \left\{ k\in\mathcal{P} :\; \mathbf{e}_{o_s}^{\top}A^h\mathbf{e}_k>0 \right\}, \\
    \Delta(\mathcal{R}_h(o_s)) &:= \left\{ \mathbf{p}\in\mathbb{R}_+^{|\mathcal{P}|} :\; \sum_{k=1}^{|\mathcal{P}|}p_k=1,\; p_k=0\ \forall k\notin\mathcal{R}_h(o_s) \right\}.
\end{aligned}
\end{equation}
where \(\mathcal{R}_h(o_s)\) denotes the set of ports that are reachable from the current origin port \(o_s\) in exactly \(h\) legs over the global liner-shipping network. Here, \(\mathbf{e}_{o_s}\) and \(\mathbf{e}_k\) are the canonical basis vectors associated with ports \(o_s\) and \(k\), respectively, and \(A^h\) is the \(h\)-th power of the adjacency matrix. Hence, the condition \(\mathbf{e}_{o_s}^{\top}A^h\mathbf{e}_k>0\) is equivalent to the existence of at least one \(h\)-step path from \(o_s\) to \(k\). Based on \(\mathcal{R}_h(o_s)\), the binary mask vector \(\mathbf{m}_{s,h}\) is introduced to indicate the admissible ports at prediction step \(h\), where \([\mathbf{m}_{s,h}]_k=1\) if and only if \(k\in\mathcal{R}_h(o_s)\). Furthermore, \(\Delta(\mathcal{R}_h(o_s))\) denotes the probability simplex restricted to the reachable set, namely, the set of all valid predictive distributions supported only on ports in \(\mathcal{R}_h(o_s)\). Therefore, the connectivity constraint is imposed not only at the level of candidate-set construction, but also directly on the support of the decoder output distribution.

Given the current origin port \(o_s\), the task is to predict an \(H\)-step future port sequence \(\mathbf{y}_s=(y_{s,1},\dots,y_{s,H})\). To enforce topological consistency with the liner-shipping network, decoding is restricted to the feasible sequence space.
\begin{equation*}
\mathcal{Y}(o_s)
:=
\left\{
\mathbf{y}=(y_1,\dots,y_H)\in\mathcal{P}^H
:\;
y_h\in \mathcal{R}_h(o_s),\ \forall h=1,\dots,H
\right\},
\end{equation*}
where \(\mathcal{R}_h(o_s)\) is defined in Eq.~\eqref{eq:reachability}. The resulting prediction is obtained via constrained maximum-probability decoding,
\begin{equation*}
\hat{\mathbf{y}}_s
\in
\operatorname*{arg\,max}_{\mathbf{y}\in\mathcal{Y}(o_s)}
p_\theta(\mathbf{y}\mid \mathcal{X}_s,\mathcal{G}),
\end{equation*}
where the joint distribution is autoregressively factorized as
\begin{equation*}
p_\theta(\mathbf{y}\mid \mathcal{X}_s,\mathcal{G})
=
\prod_{h=1}^{H}
p_\theta(y_h\mid \mathcal{X}_s,\mathbf{y}_{<h},\mathcal{G}).
\end{equation*}
Here, \(\mathcal{X}_s\) denotes the multimodal input representation, so that each step is predicted conditioned on the input, the decoded prefix, and the network structure.

\subsection{Feature Engineering}\label{sec:data-processing}
In this section, we introduce the preparation of the sample data, including the extraction of port calls and AIS features.

\subsubsection{Port-of-Call Sequence \texorpdfstring{$\bm{h_s, y_s}$}{hs, ys}.}\label{sec:voyage-segmentation}

\begin{figure}[htbp]
    \centering
    \includegraphics[width=1\linewidth, keepaspectratio]{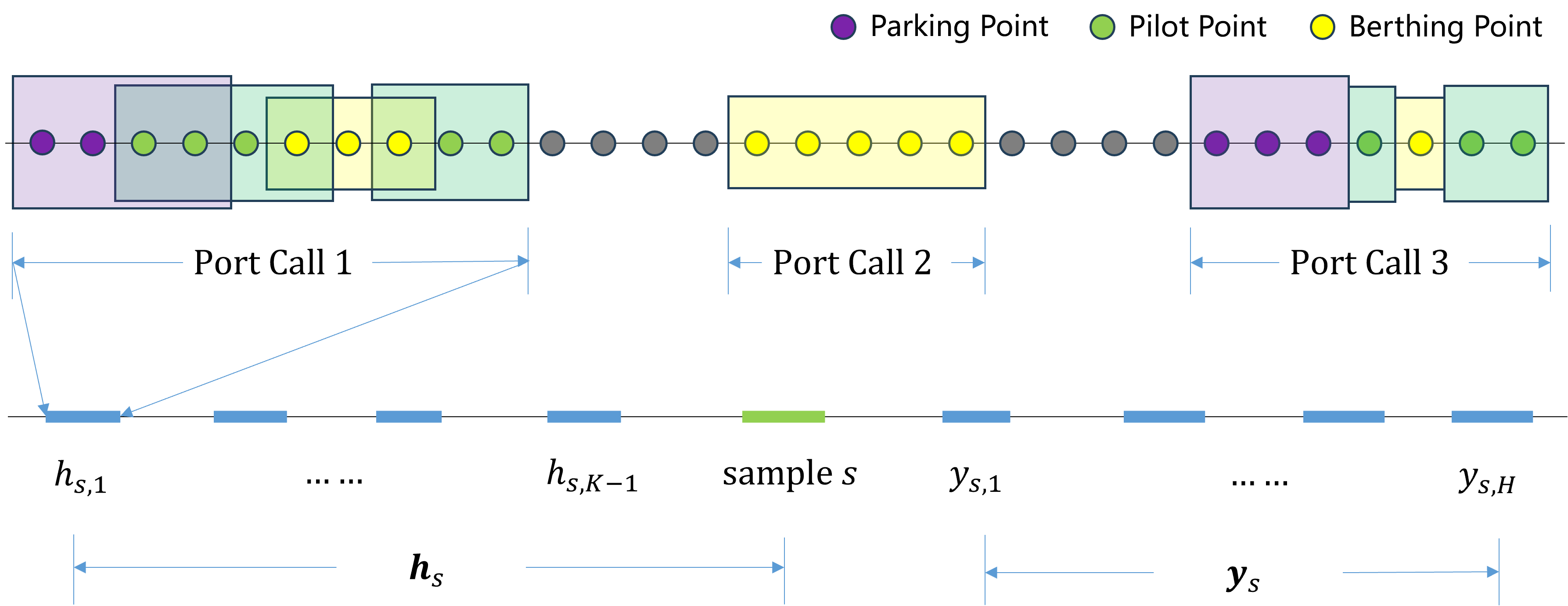}
    \caption{Voyage segmentation and sequence extraction}
    \label{fig:segments}
\end{figure}

We implement a systematic geofence-based segmentation algorithm to extract meaningful voyage segments and port-of-calls from raw AIS trajectories.
Let $\mathcal{T} =\{p_i\}_{i=1}^T= \{(lat_i, lon_i, t_i)\}_{i=1}^T$ be the raw, time-ordered AIS observations for a vessel. We define a spatial mapping function $\phi(p_i)$ that intersects each point with a set of predefined geofences encompassing berths, pilotage areas, and parking zones. We then perform sequential scanning of the temporally-ordered AIS time series to identify consecutive point segments that fall entirely within any single geofence boundary. To characterize real-world berthing dynamics more accurately and extract precise operational timestamps, we define a structured port-call sequence comprising five distinct navigation phases: parking zone entry, pilotage zone ingress, berthing arrival, post-berth pilotage re-entry, and terminal pilotage zone exit. This granular decomposition facilitates the identification of critical temporal milestones as vessels transition through different functional maritime zones. The systematic workflow for this voyage segmentation process is illustrated in Figure~\ref{fig:segments}. 

Due to missing geofence data for some ports, we cannot guarantee strict adherence to this rule for every port call. However, we can ensure strict compliance during the berthing process. Under these circumstances, these spatially-coherent segments are classified as in-port segments subject to operational validity constraints. Specifically, a segment qualifies as a genuine port stay if and only if it contains at least one berth point, thereby excluding transient passages through parking or pilotage zones that do not represent actual port calls.

Following the identification of valid in-port segments, sailing intervals are defined as the temporal complement of these port stays along the vessel's timeline, representing continuous navigation phases between distinct port calls. For each sailing interval, we establish bidirectional contextual port labels by assigning the port identifier of the preceding valid berthing segment as the previous port and the port identifier of the subsequent valid berthing segment as the next port. Specifically within the model, for any given voyage segment, we can extract its last port $h_{s,K}$. The $K-1$ ports preceding the departure port and the last port are denoted by $\bm{h}_s$, while the destination port of the voyage segment and the subsequent $H-1$ destination ports are $\bm{y_s}$.
When no subsequent port call exists within the observation window, the next port field is set to a list of $\omega$. This contextual labeling scheme will provide comprehensive port transition information for our model.

\subsubsection{Historical AIS Features $\mathcal{X}_s$.}\label{sec:feature-construction}


Following the extraction of port sequences, we construct the historical input $\mathcal{X}_s$ from AIS data. For each sample $s$, the historical AIS features are categorized into three distinct modalities: kinematic trajectory $\bm{X}^{(\mathrm{kin})}_s \in \mathbb{R}^{L_s \times d_{\mathrm{kin}}}$, vessel dynamics $\bm{X}^{(\mathrm{dyn})}_s \in \mathbb{R}^{L_s \times d_{\mathrm{dyn}}}$, and static contexts $\bm{X}^{(\mathrm{stat})}_s \in \mathbb{R}^{d_{\mathrm{stat}}}$. 
$L_s$ is the length of the AIS trajectory contained in the features; $d_{\mathrm{kin}}$, $d_{\mathrm{dyn}}$, and $d_{\mathrm{stat}}$ are the dimensions of features within the three modalities. Typically, the kinematics trajectory includes the vessel index, location and other geographical information; the vessel dynamics includes the movement features; the static contexts embody the inherent information of the vessel. To deal with 
irregular sampling in the AIS data, we use a binary mask $\bm{m}_s$ that designates $m_{s,t}=1$ for valid observations and $m_{s,t}=0$ for padded entries. For each sample $s$, we pair these feature matrices with the historical port sequence $\bm{h}_s$ and the multi-step prediction target $\mathbf{y}_s$. The specific feature entries for each component are summarized in Table~\ref{tab:feature-schema}. This comprehensive feature engineering provides a robust foundation for the multi-step port prediction architecture detailed in Section~\ref{sec:model}.

\begin{table}[htbp]
\centering
\caption{Summary of Data Variables and Feature Entries}
\label{tab:feature-schema}
\renewcommand{\arraystretch}{1.2}
\begin{tabular}{lccl}
\hline
\textbf{Category} & \textbf{Symbol} & \textbf{Dim} & \textbf{Description / Entries} \\ \hline
Kinematics trajectory & $\bm{X}^{(\mathrm{kin})}_s$ & $L_s \times d_{\mathrm{kin}}$ &  Lat, Lon, SOG, COG, Time, IMO \\
vessel dynamics & $\bm{X}^{(\mathrm{dyn})}_s$ & $L_s \times d_{\mathrm{dyn}}$ &  Speed, Draught, Heading, Course, ETA \\
Static Contexts & $\bm{X}^{(\mathrm{stat})}_s$ & $1\times d_{\mathrm{stat}}$ & Length, Width, TEU, Carrier ID, IMO \\
Validity mask & $\bm{m}_s$ & $L_s \times 1$ & Observation status (1: Valid, 0: Padding) \\ \hline
History port sequence & $\bm{h}_s$ & $K$ & Sequence of past $K$ ports ($h_{s,K} := o_s$) \\
Future port sequence & $\bm{y}_s$ & $H$ & Sequence of next $H$ ports \\ \hline
\end{tabular}
\end{table}


\subsection{The Connectivity-Constrained and Retrieval-Enhanced (CCRE) Network}\label{sec:model}
To capture the network-level information and historical patterns, we propose a CCRE neural architecture for multi-step port sequence prediction. 
\begin{figure}[htbp]
    \centering
    \includegraphics[width=1\linewidth,keepaspectratio]{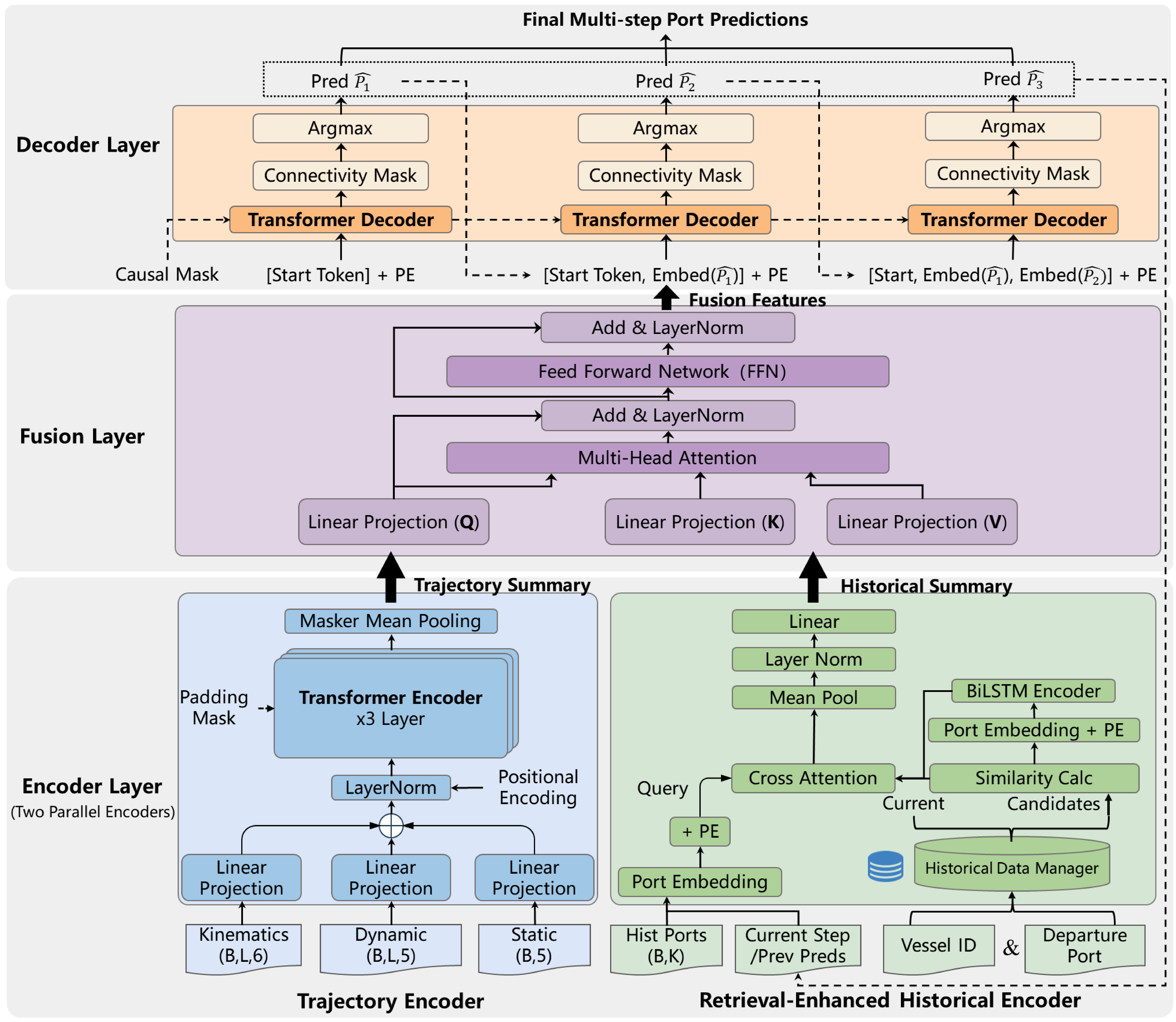}
    \caption{Overall CCRE model architecture.}
    \label{fig:model-architecture}
\end{figure}

As illustrated in Figure~\ref{fig:model-architecture}, the model comprises four synergistic components that collectively transform vessel trajectory data into feasible port sequence predictions. The process initiates with a Transformer-based trajectory encoder that distills high-resolution AIS kinematics, vessel information, and operational context into a dense latent representation. Complementing this real-time view, a retrieval-enhanced historical encoder queries a global maritime database to extract contextually relevant navigational precedents, aggregating their subsequent port patterns to capture long-term routing tendencies. To synthesize these complementary signals, an adaptive feature fusion mechanism employs cross-attention to dynamically weigh the importance of current navigational states versus retrieved historical experiences. The final stage involves a connectivity-constrained autoregressive decoder, which sequentially generates future port calls while strictly enforcing step-specific reachability constraints derived from the maritime network topology.

Formally, this forward computation pipeline transforms input features into constrained probability distributions as defined in Eq.~\eqref{eq:overall_pipeline}. The trajectory encoder $f_{\theta}^{(\mathrm{traj})}$ first maps the multiple inputs ($\bm{X}^{(\mathrm{kin})}_s, \bm{X}^{(\mathrm{dyn})}_s, \bm{X}^{(\mathrm{stat})}_s$) into a trajectory embedding $\mathbf{z}^{(\mathrm{traj})}_{s}$, while the historical encoder $g_{\theta}^{(\mathrm{hist})}$ concurrently derives a context embedding $\mathbf{z}^{(\mathrm{hist})}_{s,h}$ from retrieved precedents in $\mathcal{H}$, where $\mathcal{H}$ denotes the comprehensive repository of historical navigation scenarios indexed by origin ports and vessel identifiers to facilitate efficient retrieval. These representations are synthesized by the fusion module $\phi_{\theta}^{(\mathrm{fuse})}$ to yield a unified latent state $\mathbf{z}^{(\mathrm{fuse})}_{s,h}$. Finally, the decoder conditions on this fused representation and the step-specific topological constraint $\mathcal{R}(\cdot)$ to output the optimal port sequence $\bm{y}_s$, thereby balancing predictive accuracy with navigational feasibility.
\begin{subequations}
\label{eq:overall_pipeline}
\begin{align}
\underbrace{\mathbf{z}^{(\mathrm{traj})}_{s}}_{\text{Trajectory Encoding}}
&= f_{\theta}^{(\mathrm{traj})}\big(
    \bm{X}^{(\mathrm{kin})}_s,\,
    \bm{X}^{(\mathrm{dyn})}_s,\,
    \bm{X}^{(\mathrm{stat})}_s,\,
    \bm{m}_s
  \big), \label{eq:pipeline_traj}\\
\underbrace{\mathbf{z}^{(\mathrm{hist})}_{s,h}}_{\text{Historical Encoding}}
&= g_{\theta}^{(\mathrm{hist})}\big(
    \bm{h}_s,\,
    o_s,\,
    \hat{\bm y}_{s,<h},\,
    \mathcal{H}
  \big), \label{eq:pipeline_hist}\\
\underbrace{\mathbf{z}^{(\mathrm{fuse})}_{s,h}}_{\text{Feature Fusion}}
&= \phi_{\theta}^{(\mathrm{fuse})}\big(
    \mathbf{z}^{(\mathrm{traj})}_{s},\,
    \mathbf{z}^{(\mathrm{hist})}_{s,h}
  \big), \label{eq:pipeline_fuse}\\
y_{s,h}
&\sim
P_{\theta}
\left(
\cdot
\mid
\mathbf{z}^{(\mathrm{fuse})}_{s,h},
\hat{\bm{y}}_{s,<h},
\mathcal{R}_{h}(o_s)
\right). \label{eq:pipeline_decode}
\end{align}
\end{subequations}

This architectural framework provides a principled approach to multi-step port-call prediction that balances predictive accuracy with navigational feasibility. It leverages both current operational context and relevant historical precedents to enhance prediction quality in data-sparse maritime environments. Next, we introduce each specific network component.

\subsubsection{Transformer-Based Trajectory Encoder.}\label{sec:traj-enc}
\begin{figure}[h]
    \centering
    \includegraphics[width=0.8\linewidth,keepaspectratio]{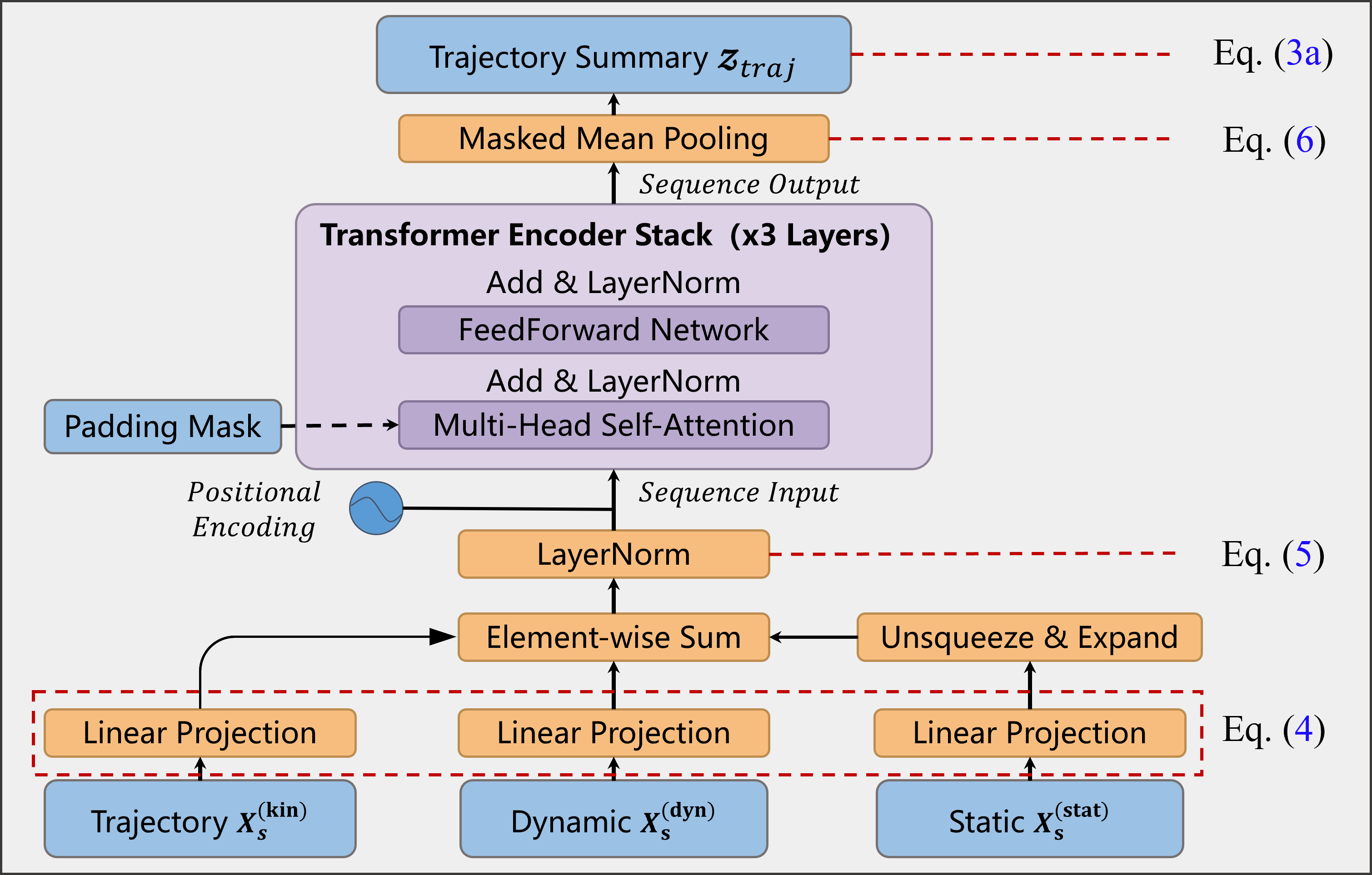}
    \caption{Trajectory Encoder.}
    \label{fig:traj-encoder-fig}
\end{figure}

As the foundational perception module illustrated in Figure~\ref{fig:traj-encoder-fig}, the trajectory encoder implements the mapping function $f_{\theta}^{(\mathrm{traj})}$ formally defined in Eq.~\eqref{eq:pipeline_traj}. It is designed to transform heterogeneous vessel navigation data into a unified representation that effectively captures both temporal dynamics and contextual operational characteristics. This module addresses the inherent challenges of integrating disparate data streams with varying temporal properties, while maintaining robustness to irregular sampling patterns.

The encoding process begins with the projection of temporally-aligned input matrices, where each feature stream is mapped to a common embedding dimension $d_{\mathrm{enc}}$ through learned linear transformations. Specifically, for each modality $k \in \{\mathrm{kin}, \mathrm{dyn}, \mathrm{stat}\}$, we compute: 
\begin{equation}
\label{eq:traj-proj}
\bm{Z}_s^{(k)} = \bm{X}_s^{(k)} \bm{W}_k + \bm{b}_k, 
\end{equation}
where $\bm{X}^{(\mathrm{kin})}_s, \bm{X}^{(\mathrm{dyn})}_s \in \mathbb{R}^{L_s \times d_k}$ represent the temporal feature matrices, $\bm{X}^{(\mathrm{stat})}_s = (\bm{X}^{(\mathrm{stat})}_s)^\top \in \mathbb{R}^{1 \times d_{\mathrm{stat}}}$ denotes the sample-level context vector, and the learnable parameters $\bm{W}_k \in \mathbb{R}^{d_k \times d_{\mathrm{enc}}}$ and $\bm{b}_k \in \mathbb{R}^{d_{\mathrm{enc}}}$ facilitate the dimensional alignment across modalities. The carrier-route projection $\bm{Z}_s^{(\mathrm{stat})}$ is subsequently broadcast along the temporal dimension to ensure consistent operational context across all time steps. 

Following the projection phase, the three feature streams undergo integration through elementwise addition, which preserves the individual contributions of each modality while enabling their synergistic combination. This fusion is stabilized through layer normalization, yielding the initial encoder representation:
\begin{equation}
\label{eq:traj-add}
\bm{U}^{(0)}_s = \text{LayerNorm}\!\big(\bm{Z}_s^{(\mathrm{kin})} + \bm{Z}_s^{(\mathrm{dyn})} + \bm{Z}_s^{(\mathrm{stat})}\big).
\end{equation}

To incorporate temporal ordering information essential for sequence modeling, standard sinusoidal positional encodings are added to $\bm{U}^{(0)}_s$, forming the input to the subsequent Transformer encoder layers. This positional encoding scheme provides the model with explicit temporal awareness while maintaining translation invariance properties.

The core contextual encoding is performed by a multi-layer Transformer encoder~\citep{Vaswani2017Attention} that processes the temporally-aligned feature sequence through masked self-attention mechanisms. These attention layers are specifically designed to handle variable-length trajectories and missing observations by utilizing padding masks derived from the validity mask $\bm{m}_s$. This masking strategy ensures that invalid positions do not contribute to attention computations, thereby preventing the propagation of spurious information through the network. The resulting contextualized sequence representation $\bm{U}_s \in \mathbb{R}^{L_s \times d_{\mathrm{enc}}}$ captures both fine-grained local temporal patterns and broader global trajectory characteristics, providing a comprehensive encoding of the vessel's navigational behavior.

To obtain a fixed-dimensional representation suitable for downstream processing modules, the encoder applies masked mean pooling over the valid time steps. This aggregation operation computes:
\begin{equation}
\label{eq:traj-pool}
\mathbf{z}^{(\mathrm{traj})}_{s} = \frac{\sum_{t=1}^{L_s} m_{s,t} \cdot \bm{U}_{s,t}}{\sum_{t=1}^{L_s} m_{s,t}} \in \mathbb{R}^{d_{\mathrm{enc}}},
\end{equation}
where $m_{s,t}$ denotes the validity indicator for time step $t$ in sample $s$, and $\bm{U}_{s,t}$ represents the $t$-th row of the contextualized sequence representation. This trajectory summary $\mathbf{z}^{(\mathrm{traj})}_{s}$ serves as the primary conditioning input to subsequent retrieval-fusion and decoding modules, encapsulating the essential navigational characteristics of the current voyage segment in a compact yet informative representation.

The trajectory encoder thus establishes a principled framework for multiple feature integration that maintains temporal coherence while accommodating the irregular sampling patterns and data quality variations inherent in real-world maritime AIS data. This robust encoding foundation enables the downstream components to effectively leverage the rich temporal and contextual information contained within vessel trajectories for accurate port sequence prediction.

\subsubsection{Retrieval-Enhanced Historical Encoder.}\label{sec:hist-enc}

In the domain of global liner shipping, conventional sequence learning models face two fundamental impediments: the long-tail sparsity of port-call distributions and the heterogeneity of vessel routing behavior patterns. Specifically, purely kinematic-driven approaches struggle to generalize on infrequent routes due to insufficient training supervision, while simultaneously failing to disambiguate complex decision-making scenarios where similar physical states may lead to divergent destinations on the whole maritime network.

\begin{figure}[h]
    \centering
    \includegraphics[width=0.9\linewidth,keepaspectratio]{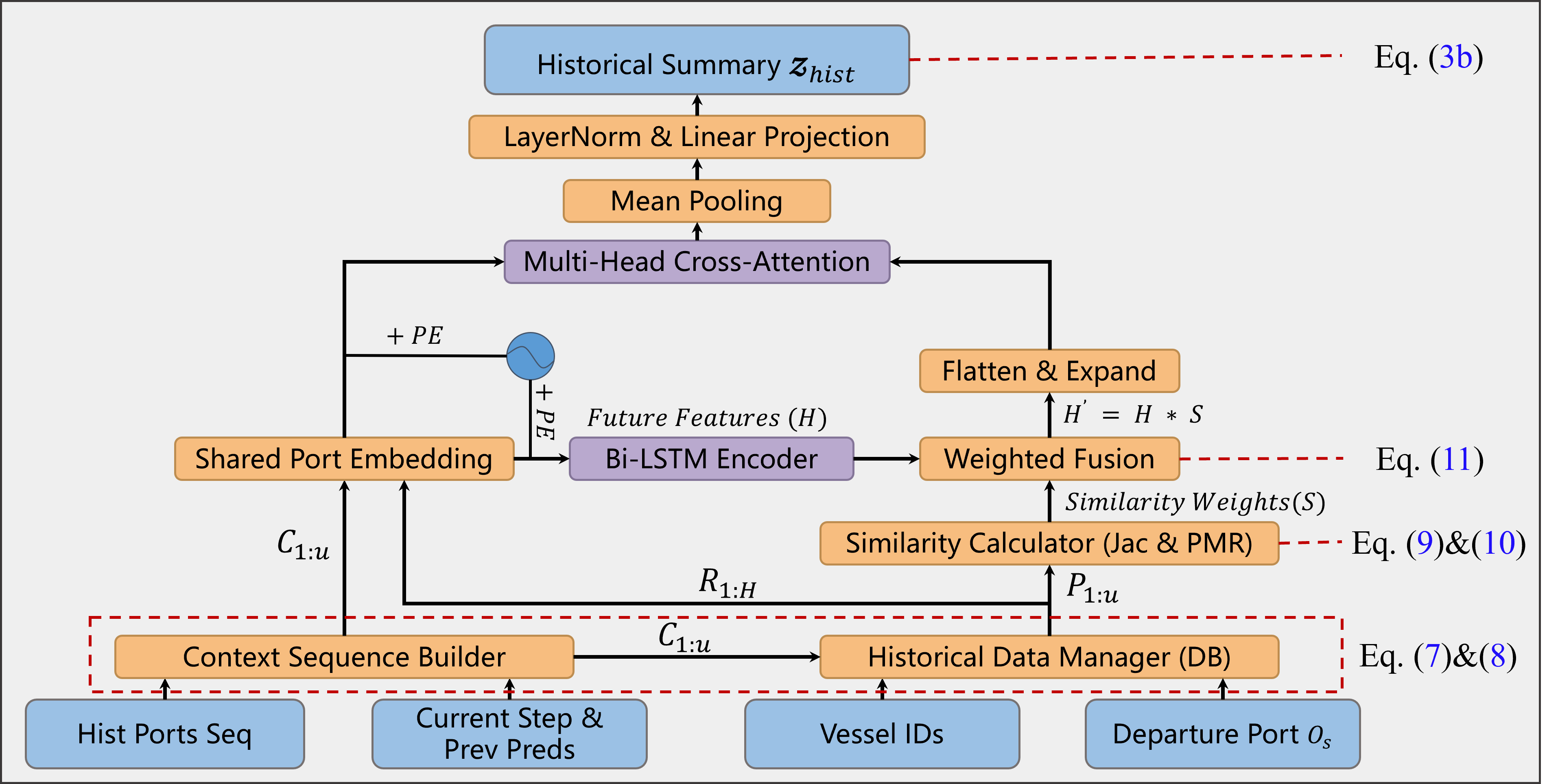}
    \caption{Historical Encoder.}
    \label{fig:hist-encoder-fig}
\end{figure}

To bridge these gaps, we introduce the retrieval-enhanced historical encoder, illustrated in Figure~\ref{fig:hist-encoder-fig}, which implements the mapping function $g_{\theta}^{(\mathrm{hist})}$ formally defined in Eq.~\eqref{eq:pipeline_hist}. Motivated by the intuition that historical precedents serve as strong inductive priors, this module explicitly queries and aggregates contextually similar past voyages from a global database. By conditioning predictions on retrieved historical patterns rather than relying solely on instantaneous kinematic states, our approach effectively compensates for data scarcity on rare routes and resolves routing ambiguities, thereby grounding the model in established operational logic.

Formally, we consider the retrieval process at prediction step $h$, where the model has generated a partial sequence. We define the effective context length as $u = K + h - 1$, which represents the summation of the observed history length $K$ and the $h-1$ generated steps.
To retrieve the most relevant scenarios, we construct a dynamic query context $C_{1:u}$ by concatenating the vessel's input history and the ports predicted thus far:
\begin{equation}
\label{eq:dynamic-context}
C_{1:u} = \bm{h_{s}} \oplus \hat{y}_{s, <h} \in (\mathcal{P}\cup\{\omega\})^{u},
\end{equation}
where $\bm{h_s}$ is the observed historical port sequence defined in Eq.~\eqref{eq:history_sequence}, $\hat{y}_{s, <h}$ denotes the generated sequence, and $\oplus$ represents the concatenation operator.

Correspondingly, we establish a historical database $\mathcal{H}$ indexed by the origin port. For each historical scenario $\mathcal{S}_n \in \mathcal{H}$, we extract a prefix sequence aligned with the current context length $u$. Thus, the $n$-th scenario utilized for comparison is formulated as:
\begin{equation}
\label{eq:historical-scenarios}
\mathcal{S}_n = \big(P^{(n)}_{1:u}, R^{(n)}_{1:H}\big), \qquad
P^{(n)}_{1:u} \in (\mathcal{P}\cup\{\omega\})^{u},\quad
R^{(n)}_{1:H} \in (\mathcal{P}\cup\{\omega\})^{H},
\end{equation}
where $P^{(n)}_{1:u}$ denotes the historical prefix of length $u$ (implicitly truncated or padded from the raw record to match the query), and $R^{(n)}_{1:H}$ represents the ground-truth subsequent $H$ ports serving as the prediction target.

To quantify the relevance of each historical scenario $\mathcal{S}_n$, we employ two complementary similarity metrics between the query $C_{1:u}$ and the historical prefix $P^{(n)}_{1:u}$.
First, to capture global routing intent, we utilize the Jaccard Similarity~\citep{gusfield1997algorithms}, which measures the set-level overlap regardless of temporal order:
\begin{equation}
\mathrm{Jac}(P^{(n)}_{1:u}, C_{1:u})
=
\frac{
|\mathcal{U}_{\omega}(P^{(n)}_{1:u}) \cap \mathcal{U}_{\omega}(C_{1:u})|
}{
|\mathcal{U}_{\omega}(P^{(n)}_{1:u}) \cup \mathcal{U}_{\omega}(C_{1:u})|
},
\end{equation}
where $\mathcal{U}_{\omega}(\cdot)$ denotes the set of unique valid ports after removing the sentinel token $\omega$.
Second, to ensure precise trajectory alignment, we introduce the Positional Match Rate (PMR), inspired by the exact matching criteria in sequence alignment methods~\citep{choi2010survey}. Since both sequences share the same effective length $u$, this metric strictly evaluates the sequential agreement of the trajectory:
\begin{equation}
\mathrm{PMR}(P^{(n)}_{1:u}, C_{1:u})
=
\frac{
\sum_{t=1}^{u}
\mathbb{I}
\left(
P^{(n)}_{t}=C_t,\;
P^{(n)}_{t}\neq\omega,\;
C_t\neq\omega
\right)
}{
\max\left\{
1,\;
\sum_{t=1}^{u}
\mathbb{I}
\left(
P^{(n)}_{t}\neq\omega,\;
C_t\neq\omega
\right)
\right\}
}.
\end{equation}
where $\mathbb{I}(\cdot)$ is the indicator function. The PMR prioritizes scenarios where the vessel followed an identical path immediately preceding the current state.

The final relevance score $s_n$ is a weighted combination of these metrics: $s_{s,h}^{(n)} = \alpha \cdot \mathrm{Jac} + (1-\alpha) \cdot \mathrm{PMR}$, where $\alpha\in[0,1]$ controls the relative importance of set-level route similarity and position-wise sequential alignment. For computational efficiency and to avoid introducing noisy precedents, we retain the top-$N$ historical scenarios according to $s_{s,h}^{(n)}$. Let $\mathcal{I}_{s,h} = \operatorname{TopN}_{n:\mathcal{S}_n\in\mathcal{H}(o_s)}
\left\{
s_{s,h}^{(n)}
\right\}$ denote the index set of the retrieved scenarios, where $\mathcal{H}(o_s)$ denotes the subset of the historical database associated with the current origin port $o_s$. The similarity scores of the retrieved scenarios are then normalized into retrieval-prior weights:
\begin{equation*}
\label{eq:retrieval-prior}
w_{s,h}^{(n)}
=
\frac{
\exp\left(s_{s,h}^{(n)}/\tau_r\right)
}{
\sum_{m\in\mathcal{I}_{s,h}}
\exp\left(s_{s,h}^{(m)}/\tau_r\right)
},
\qquad n\in\mathcal{I}_{s,h},
\end{equation*}
where $\tau_r>0$ is a temperature parameter controlling the sharpness of the retrieval distribution. These weights are used for selecting the top-N retrieved scenarios.
We then encode the future component $R^{(n)}_{1:H}$ using a Bidirectional LSTM (BiLSTM)~\citep{graves2005framewise} to capture temporal dynamics.

Instead of averaging temporal steps which may dilute sequential dependencies, we extract the final hidden states from both directions to represent the semantic summary of the trajectory. Let $\mathbf{h}^{(n)}_{s,h} = \left[
\overrightarrow{\mathbf{h}}^{(n)}_H;
\overleftarrow{\mathbf{h}}^{(n)}_1
\right]\mathbf{W}_{r}+
\mathbf{b}_{r}$ denotes the semantic embedding of the $n$-th retrieved continuation. The retrieved scenario representations are then stacked into a candidate-level historical representation:
\begin{equation}
\label{eq:final-hist-agg}
\mathbf{z}^{(\mathrm{hist})}_{s,h}
=
\begin{bmatrix}
\left(\mathbf{h}_{s,h}^{(n_1)}\right)^\top \\
\left(\mathbf{h}_{s,h}^{(n_2)}\right)^\top \\
\vdots \\
\left(\mathbf{h}_{s,h}^{(n_N)}\right)^\top
\end{bmatrix}
\in\mathbb{R}^{N\times d_r},
\qquad
n_i\in\mathcal{I}_{s,h},
\end{equation}

Here, $\mathbf{z}^{(\mathrm{hist})}_{s,h}$ preserves the retrieved historical continuations as separate candidate-level embeddings.

To accommodate the autoregressive nature of multi-step prediction, we employ a query-driven decoding strategy. Unlike static approaches, our mechanism iteratively refines the utilization of the encoded context at each decoding step. Specifically, as new port predictions are generated, the model updates its internal decoder state by incorporating the latest output. This evolved state serves as the dynamic query to compute cross-attention weights over the retrieved historical candidate representations. This step-wise attention shift ensures that the model actively focuses on the specific historical patterns most relevant to the current prediction stage, allowing it to adaptively exploit routing precedents that align with the unfolding trajectory.

\subsubsection{Adaptive Feature Fusion.}\label{sec:fusion}
The feature fusion module defined in Eq.~\eqref{eq:pipeline_fuse} serves as the central integration point at which trajectory-based navigation patterns and similarity-retrieved historical contexts are combined into a unified representation for multi-step prediction. Its objective is to balance the short-term dynamics captured by the trajectory encoder with the longer-term routing regularities encoded in the historical module, while maintaining computational efficiency and interpretability. Similar multi-modal integration strategies have demonstrated robustness in recent maritime trajectory prediction studies~\citep{Yu2025multi, luo2025multimodal}.

Given the trajectory summary obtained from the encoder in Eq.~\eqref{eq:traj-pool} and the historical context matrix produced by the similarity-based historical encoder in Eq.~\eqref{eq:final-hist-agg}, the fusion block first projects the trajectory vector and the historical candidate matrix into a common fusion space of dimension $(d_{\mathrm{fuse}})$ via learned affine transformations following the standard attention mechanism architecture~\citep{Vaswani2017Attention}:
\begin{align*}
\mathbf{Q}_{s,h}&=\mathbf{z}^{(\mathrm{traj})}_{s}\mathbf{W}_{Q}
+\mathbf{b}_{Q},&\mathbf{W}_{Q}&\in\mathbb{R}^{d_{\mathrm{enc}}\times d_{\mathrm{fuse}}},
\\
\mathbf{K}_{s,h}&=\mathbf{z}^{(\mathrm{hist})}_{s,h}\mathbf{W}_{K}
+\mathbf{b}_{K},&\mathbf{W}_{K}&\in\mathbb{R}^{d_r\times d_{\mathrm{fuse}}},
\\
\mathbf{V}_{s,h}&=\mathbf{z}^{(\mathrm{hist})}_{s,h}\mathbf{W}_{V}
+\mathbf{b}_{V},&\mathbf{W}_{V}&\in\mathbb{R}^{d_r\times d_{\mathrm{fuse}}}.
\end{align*}
Here, \(\mathbf{Q}_{s,h}\in\mathbb{R}^{1\times d_{\mathrm{fuse}}}\) is a single-query representation derived from the current voyage, whereas \(\mathbf{K}_{s,h},\mathbf{V}_{s,h}\in\mathbb{R}^{N\times d_{\mathrm{fuse}}}\) are key-value representations of the \(N\) retrieved historical candidates. This asymmetric design ensures that the trajectory encoder retains semantic precedence, while historical information is incorporated only when deemed relevant by the attention mechanism.

The core fusion operation is implemented as a cross-attention layer in the fusion space. We compute the attention output $\mathbf{O}^{\mathrm{attn}}_{s,h}$ using the scaled dot-product mechanism~\citep{Vaswani2017Attention}: 
\begin{equation*}
\label{eq:attn-calc}
\mathbf{O}^{\mathrm{attn}}_{s,h} = \text{softmax}\left(
\frac{
\mathbf{Q}_{s,h}\mathbf{K}_{s,h}^{\top}
}{
\sqrt{d_{\mathrm{fuse}}}
}\right)\mathbf{V}_{s,h},
\end{equation*}
where the softmax is applied along the last dimension. This operation enables the model to selectively attend to historical scenarios that are most compatible with the current navigational state, while suppressing irrelevant or conflicting patterns. The resulting attention weights act as a soft, data-dependent candidate-selection mechanism that modulates which retrieved historical precedents contribute more strongly to the fused representation.

The attention output is combined with the query through residual connections~\citep{he2016deep} and layer normalization~\citep{ba2016layer}, followed by a position-wise feed-forward network. Denoting by $\mathrm{Drop}(\cdot)$ a dropout layer and by $\mathrm{LN}_1,\mathrm{LN}_2$ two layer-normalization operators, the fusion block computes
\begin{align*}
\widetilde{\mathbf{z}}_{s,h}
&=
\operatorname{LN}_{1}
\left(
\mathbf{Q}_{s,h}
+
\operatorname{Drop}
\left(
\mathbf{O}^{\mathrm{attn}}_{s,h}
\right)
\right).\\
\mathbf{z}^{(\mathrm{fuse})}_{s,h}
&=
\operatorname{LN}_{2}
\left(
\widetilde{\mathbf{z}}_{s,h}
+
\operatorname{Drop}
\left(
\operatorname{FFN}
\left(
\widetilde{\mathbf{z}}_{s,h}
\right)
\right)
\right).
\end{align*}
where $\mathbf{z}^{(\mathrm{fuse})}_{s,h} \in \mathbb{R}^{1 \times d_{\mathrm{fuse}}}$ is the final fused representation. The feed-forward network adopts the standard two-layer Transformer structure,
\begin{equation*}
\mathrm{FFN}(\mathbf{x}) 
= \phi\!\big(\mathbf{x}\mathbf{W}_1 + \mathbf{b}_1\big)\mathbf{W}_2 + \mathbf{b}_2,
\end{equation*}
with $\mathbf{W}_1 \in \mathbb{R}^{d_{\mathrm{fuse}}\times d_{\mathrm{ff}}}$, $\mathbf{W}_2 \in \mathbb{R}^{d_{\mathrm{ff}}\times d_{\mathrm{fuse}}}$, $\mathbf{b}_1 \in \mathbb{R}^{d_{\mathrm{ff}}}$, $\mathbf{b}_2 \in \mathbb{R}^{d_{\mathrm{fuse}}}$, and nonlinearity $\phi(\cdot)$ instantiated as a ReLU activation.

This design yields a trajectory-centered yet history-aware representation: the query path $\mathbf{z}^{\mathrm{traj}}_s$ is preserved through residual connections, while the cross-attention and feed-forward sublayers inject information about historically observed continuation patterns in a similarity-dependent manner. In the multi-step autoregressive setting, the historical encoder is re-evaluated as new predictions are generated, and the fusion block is applied at each decoding step with an updated context. Consequently, $\mathbf{z}^{(\mathrm{fuse})}_{s,h}$ provides a dynamically adapted summary that jointly reflects the current navigational scenario and the most relevant routing precedents, and serves as the input to the downstream port prediction module.

\subsubsection{Connectivity-Constrained Autoregressive Decoder.}\label{sec:decoder}

We formulate multi-step port forecasting as an autoregressive sequence modeling problem. 
Conditioned on a linear projection of the fused representation $\mathbf{z}^{(\mathrm{fuse})}_{s,h}$ as decoder memory, a $D$-layer Transformer decoder~\citep{Vaswani2017Attention} generates the $H$ future ports in an auto-regressive manner. At each decoding step, the input sequence consists of a learnable Beginning Of Sequence (\emph{BOS}) token followed by the embeddings of previously predicted ports, and a standard upper-triangular causal mask is applied to preserve auto-regressivity. 
During training, teacher forcing is used to condition the decoder on ground-truth prefixes, and we optionally employ the Gumbel-Softmax relaxation~\citep{jang2016categorical} (with temperature $\tau=1.0$) to obtain differentiable hard samples of intermediate port predictions. Specifically, the probability of sampling a port $i$ at step $h$ is formulated as:
\begin{equation*}\label{eq:gumbel_softmax}
    \hat{y}_{s,h}^{(i)} = \frac{\exp\left((\tilde{l}_{s,h}(i) + g_i) / \tau\right)}{\sum_{j \in \mathcal{R}_h(o_s)} \exp\left((\tilde{l}_{s,h}(j) + g_j) / \tau\right)},
\end{equation*}
where $g_i \sim \text{Gumbel}(0,1)$ are independent and identically distributed samples drawn from the Gumbel distribution, and $\tau$ is the temperature parameter controlling the sharpness of the sampling distribution. At inference time, predictions are obtained by greedy decoding via $\arg\max$ over the constrained output distribution at each step.

Given the multiple inputs and historical context, the predictive architecture factorizes the joint distribution of the $H$-step target sequence $\bm{y}_s = (y_{s,1},\ldots,y_{s,H})$ as
\begin{equation}
\label{eq:chain}
p_{\theta}
\left(
\bm{y}_{s}
\mid
\mathbf{X}^{(\mathrm{kin})}_{s},
\mathbf{X}^{(\mathrm{dyn})}_{s},
\mathbf{X}^{(\mathrm{stat})}_{s},
\bm{h}_{s},
o_s
\right)
=
\prod_{h=1}^{H}
p_{\theta}
\left(
y_{s,h}
\mid
\mathbf{z}^{(\mathrm{fuse})}_{s,h},
\hat{\bm{y}}_{s,<h}
\right).
\end{equation}
where $\mathbf{z}^{(\mathrm{fuse})}_{s,h}$ denotes the fused representation produced by the trajectory encoder, the similarity-based historical encoder, and the fusion module, and $\hat{\bm{y}}_{s,<h}$ denotes the sequence of ports predicted at previous steps $1,\ldots,h-1$. This autoregressive factorization enables sequential decision-making while maintaining a coherent probabilistic semantics over the entire prediction horizon.
\begin{figure}[h]
    \centering
    \includegraphics[width=0.8\linewidth,keepaspectratio]{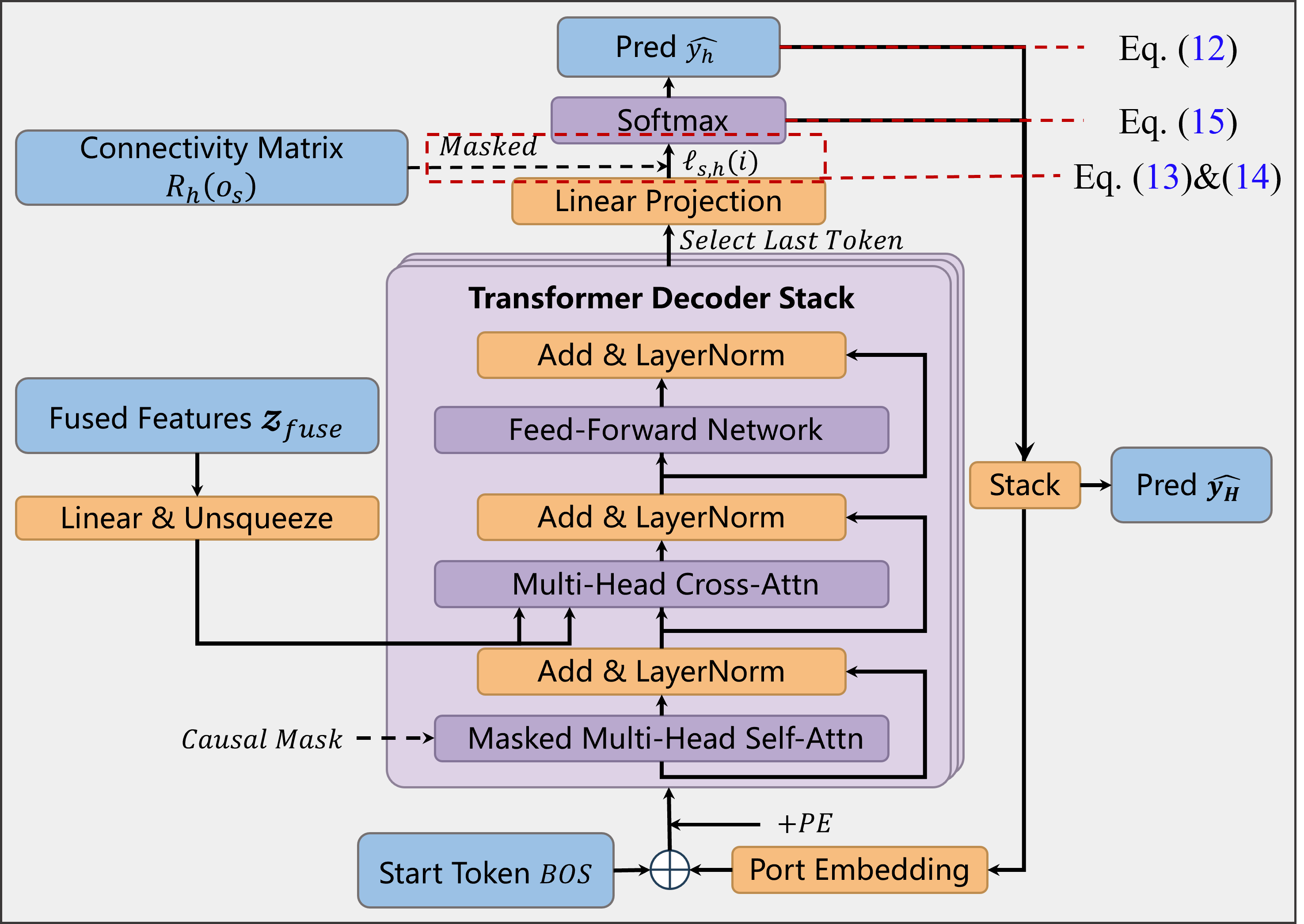}
    \caption{Auto-regressive Decoder.}
    \label{fig:decoder-fig}
\end{figure}

To ensure that the generated routes are navigationally feasible, we impose hard topological constraints derived from the historical port-to-port connectivity network structure $\mathcal{G}$ defined in Section~\ref{sec:formulation}. Instead of allowing unconstrained generation over the entire vocabulary $\mathcal{P}$, we restrict the decoder's output space at each step. Specifically, for a given origin port $o_s$, the valid candidate set for the next prediction step is strictly bounded by the feasibility domain $\mathcal{R}(o_s)\subseteq \mathcal{P}$, which effectively prunes operationally impossible connections from the search space.
This is enforced by restricting the support of each step-wise predictive distribution to the corresponding reachability set:
\begin{equation}
\label{eq:reachability-constrain}
\operatorname{supp}\!\big(p_{\theta}(y_{s,h}\mid\cdot)\big) \subseteq \mathcal{R}_h(o_s),
\qquad h = 1,\ldots,H.
\end{equation}

In practice, let $\ell_{s,h}(i)$ denote the unconstrained logit assigned by the decoder to a candidate port index $i \in \mathcal{P}$ at prediction step $h$. A step-dependent Boolean mask is applied to enforce the reachability constraint: 
\begin{equation}
\label{eq:masked-logits}
\tilde{\ell}_{s,h}(i) =
\begin{cases}
\ell_{s,h}(i), & i \in \mathcal{R}_h(o_s),\\[2pt]
-\infty, & i \notin \mathcal{R}_h(o_s),
\end{cases}
\end{equation}
and the resulting masked logits are normalized with a softmax over the feasible set. The probability of selecting port $i$ is given by: 
\begin{equation}
\label{eq:masked-softmax}
p_{\theta}\big(y_{s,h} = i \mid \mathbf{z}^{(\mathrm{fuse})}_{s,h},\hat{\bm{y}}_{s,<h}\big)
= \frac{\exp\big(\tilde{\ell}_{s,h}(i)\big)}{\displaystyle\sum_{j \in \mathcal{R}_h(o_s)} \exp\big(\tilde{\ell}_{s,h}(j)\big)}. 
\end{equation}
Equivalently, $p_{\theta}(y_{s,h}=i\mid\cdot)=0$ for all $i \notin \mathcal{R}_h(o_s)$, so the per-step support of the predictive distribution is always a subset of the connectivity-induced feasible set.

This connectivity-constrained auto-regressive decoder thus couples a flexible neural sequence model with an explicit graph-based feasibility prior. The fused context $\mathbf{z}^{(\mathrm{fuse})}_{s,h}$ ensures that each decision reflects both the current navigational state and the most relevant historical routing precedents, while the reachability masks guarantee that the resulting multi-step port sequences remain consistent with the underlying maritime network topology.

\subsubsection{Learning Objective.}\label{sec:learning-objective}
We formulate parameter learning as a constrained optimization problem that minimizes prediction error while enforcing consistency with the maritime connectivity structure. The objective further incorporates label smoothing~\citep{szegedy2016rethinking} for improved probabilistic calibration and robust handling of missing supervision signals that frequently arise in real-world trajectory data.

In our connectivity-constrained framework, we propose a connectivity-constrained variant of label smoothing. Unlike standard approaches that distribute probability mass uniformly across the entire vocabulary, our method redistributes the smoothing mass $\varepsilon$ exclusively among the navigationally feasible ports defined by the reachability set $\mathcal{R}_h(o_s)$ in Eq.~\eqref{eq:reachability}.
Formally, the connectivity-constrained label-smoothed loss $\mathcal{L}_{\epsilon}$ for a single prediction step $h$ is defined as:
\begin{equation*}\label{eq:label_smoothing}
    \mathcal{L}_{\epsilon}(p_\theta, y_{s,h}) = - (1 - \epsilon) \log p_\theta(y_{s,h} \mid \mathbf{z}^{(\mathrm{fuse})}_{s,h}, \hat{\mathbf{y}}_{s,<h}) - \frac{\epsilon}{|\mathcal{R}_h(o_s)|} \sum_{j \in \mathcal{R}_h(o_s)} \log p_\theta(j \mid \mathbf{z}^{(\mathrm{fuse})}_{s,h}, \hat{\mathbf{y}}_{s,<h}),
\end{equation*}
where $\epsilon\in[0,1]$ controls the degree of smoothing. This formulation redistributes a fraction $\epsilon$ of the target mass from the ground-truth port to all feasible alternatives within $\mathcal{R}_{h}(o_{s})$, thereby improving calibration and mitigating overconfidence while remaining consistent with connectivity constraints.

Building upon this regularized loss, we formulate the parameter learning process as a constrained optimization problem. The goal is to minimize the cumulative error over the training set $\mathcal{D}_{\mathrm{train}}$ while strictly adhering to the topological boundaries. The global learning objective is formulated as follows:
\begin{equation}
\label{eq:objective}
\begin{aligned}
\min_{\theta}\;\;
&\sum_{s\in\mathcal{D}_{\mathrm{train}}}\sum_{h=1}^{H} 
\mathbb{I}\{y_{s,h} \neq \omega\} \,
\mathcal{L}_{\varepsilon}\!\Big(
  p_{\theta}(\cdot \mid \mathbf{z}^{(\mathrm{fuse})}_{s,h},  \hat{y}_{s,<h}), 
  \,y_{s,h}
\Big) \\
\text{s.t.}\;\;
&\operatorname{supp}\!\big(
  p_{\theta}(\cdot \mid \mathbf{z}^{(\mathrm{fuse})}_{s,h},  \hat{y}_{s,<h}) 
\big)
\subseteq \mathcal{R}_h(o_s),
\qquad h = 1,\ldots,H, 
\end{aligned}
\end{equation}
where $\mathbb{I}\{\cdot\}$ denotes the indicator function that excludes positions with missing or sentinel labels ($y_{s,h} = \omega$) from the loss, and the constraint enforces that the support of the model distribution is contained within the step-dependent reachability set $\mathcal{R}_h(o_s)$.


Constraint enforcement is implemented via the hard masking mechanism in Eq.~\eqref{eq:masked-logits}, which sets the logits of all ports outside $\mathcal{R}_h(o_s)$ to $-\infty$ prior to normalization. As a result, $p_{\theta}(y_{s,h}=i\mid\cdot)=0$ for all $i\notin\mathcal{R}_h(o_s)$, and the constraint in \eqref{eq:objective} is satisfied by construction. This masking is applied uniformly during both training and inference, ensuring that the model can only allocate probability mass to navigationally feasible ports while preserving differentiability for gradient-based optimization. 

The learning objective simultaneously promotes predictive accuracy, improves probabilistic calibration, enhances robustness to incomplete supervision, and enforces operational feasibility through connectivity constraints. Taken together, these elements define a training framework that is explicitly tailored to the structural properties and data quality characteristics of maritime route prediction.
\section{Experiment Design}\label{sec:design}
This section outlines the experimental setup employed to evaluate the proposed framework. We detail the construction and statistical characteristics of the global-scale container shipping dataset, define the evaluation metrics used to assess multi-step prediction performance, and describe the baseline configurations designed to benchmark the proposed approach.
\subsection{Datasets and Statistics}\label{sec:data}
The primary dataset has been constructed from two distinct sources, which have undergone systematic preprocessing to construct a comprehensive maritime voyage prediction benchmark.

The first source comprises raw AIS trajectory data, encompassing voyage records for a multitude of maritime vessels throughout the year 2021. As formally defined in Table~\ref{tab:ais-schema}, this data furnishes high-resolution spatiotemporal information, incorporating vessel positions, velocities, headings, and timestamps at regular intervals. To guarantee the caliber and pertinence of the dataset for global liner shipping route prediction, we implemented a stringent, three-stage vessel filtering mechanism on the raw AIS dataset, which initially contained over 5,000 maritime targets. First, we applied strict vessel classification code filtering to isolate container ships, systematically excluding non-liner vessels such as bulk carriers, oil tankers, and fishing vessels. Second, to ensure operational continuity, we eliminated vessels with severe AIS signal sparsity, extensive transmission gaps, or prolonged laid-up periods during 2021. Finally, we excluded localized feeder vessels and inland barges whose trajectories were strictly confined to single-country coastal or riverine areas, as they do not participate in the international liner network. This rigorous pipeline yielded a highly curated cohort of 400 distinct, actively trading deep-sea container vessels identified by unique International Maritime Organization (IMO) numbers.

Although this cohort represents a subset of the global fleet, it is exceptionally representative of the global liner shipping network topology. Driven by the hub-and-spoke nature of global maritime trade, this 400-vessel fleet generated 29,097 complete voyage segments over the year. More importantly, their aggregated operational footprint spans 602 unique global ports and explicitly captures 3,867 distinct port-to-port connectivity edges—encompassing all major intercontinental trade lanes, including the Trans-Pacific, Asia-Europe, and Trans-Atlantic routes. Therefore, the routing behaviors extracted from this cohort provide a dense, statistically significant, and geographically comprehensive foundation for modeling multi-step predictive logic on a truly global scale.

\begin{table}[h]
  \centering
  \caption{Schema of Raw AIS Data}
  \label{tab:ais-schema}
  \begin{tabularx}{\linewidth}{l l l}
    \toprule
    \textbf{Field Name} & \textbf{Data Type} & \textbf{Description} \\
    \midrule
    \texttt{IMO} & String & International Maritime Organization number \\
    \texttt{timestamp} & String & Timestamp of the event generation \\
    \texttt{latitude} & Double & Latitude coordinate of the vessel \\
    \texttt{longitude} & Double & Longitude coordinate of the vessel \\
    \texttt{speed} & Double & Speed over ground (in knots) \\
    \texttt{course} & Double & Course over ground (intended direction) \\
    \texttt{heading} & Double & True heading (direction of the bow) \\
    \texttt{draught} & Double & Vertical distance from waterline to keel \\
    \texttt{destination} & String & Destination reported by the captain \\
    \texttt{ETA} & String & estimated time of arrival at destination \\
    \bottomrule
    \end{tabularx}
\end{table}

The second dataset comprises geospatially referenced port infrastructure data, structured as outlined in Table~\ref{tab:geofence-schema}. This dataset includes precisely delineated geofenced boundaries for global port berthing zones, pilotage areas, and individual berth locations. This geospatial reference framework enables the systematic segmentation of continuous vessel trajectories into discrete voyage segments, where each segment represents a complete journey between two consecutive port visits. By applying the geofence-based segmentation methodology detailed in Section~\ref{sec:voyage-segmentation}, we identified 29,097 individual voyage segments across the 400-vessel fleet, with each vessel contributing between 7 and 295 segments.

\begin{table}[h]
  \centering
  \caption{Schema of Port Geofence Data}
  \label{tab:geofence-schema}
  \begin{tabular}{l l l}
    \toprule
    \textbf{Field Name} & \textbf{Data Type} & \textbf{Description} \\
    \midrule
    \texttt{portId} & Integer & Unique identifier for the port area \\
    \texttt{geometry} & String & Spatial polygon coordinates in WKT format \\
    \texttt{polygonType} & String & Functional classification (e.g., Parking zone, Pilot zone) \\
    \bottomrule
  \end{tabular}
\end{table}

For each of these identified segments, we extract multi-dimensional feature vectors that capture both static vessel characteristics and dynamic operational parameters, strictly adhering to the multiple feature schema defined in Section~\ref{sec:feature-construction}. Subsequent to the temporal resampling and spatial filtering, the final dataset comprises over 5.26 million georeferenced observations. The processed voyage segments exhibit substantial variability in duration, with sequence lengths ranging from 2 to 3,833 points (mean: 180.8). Each time step within these sequences is fully annotated with the aligned kinematics $\bm{X}^{(\mathrm{kin})}_s$ and vessel dynamics features $\bm{X}^{(\mathrm{dyn})}_s$.

Static vessel characteristics are encapsulated in the static context vector $\bm{X}^{(\mathrm{stat})}_s$ derived from the Carrier (CRR), covering essential attributes such as dimensions, Twenty-foot Equivalent Unit (TEU) capacity, and unique identifiers. To configure the multi-step sequence modeling task, we standardize the historical context window to $K=3$ (comprising the three most recent port visits) and set the prediction horizon to $H=3$ (targeting the subsequent three destination ports). The finalized dataset structure, strictly aligned with the schema in Table~\ref{tab:data-schema}, serves as the standardized input for model training and evaluation.

\begin{table}[h]
  \centering
  \small
  \caption{Segment-level data schema.}
  \label{tab:data-schema}
  \begin{tabularx}{\textwidth}{l l l X}
    \toprule
    \textbf{Field Name} & \textbf{Data Type}  & \textbf{Description} & \textbf{Example Value} \\
    \midrule
    \texttt{imo}                 & Integer      & Unique Ship Identifier  & \texttt{9863895} \\
    \texttt{start\_time}         & Timestamp    & Voyage segment start time   & \texttt{2021-01-10 13:00:00} \\
    \texttt{end\_time}           & Timestamp    & Voyage segment end time  & \texttt{2021-01-15 08:30:00} \\
    \texttt{origin}              & String       & Departure port name    & \texttt{"Singapore"} \\
    \texttt{destination}         & String       & Arrival port name     & \texttt{"Surabaya"} \\
    \texttt{feature\_kin}       & Float64 array & Kinematics trajectories&\([\text{lat}, \text{lon}, \text{speed}, \text{course}, \text{time}, \text{imo}]\)   \\
    \texttt{feature\_dyn}     & List of Tuples  & Dynamic vessel features&\([\text{speed}, \text{draft}, \text{heading}, \text{course}, \text{time}]\)   \\
    \texttt{feature\_stat}        & Tuple     & static carrier features&\([\text{length}, \text{width}, \text{TEU}, \text{crrId}, \text{imo}]\)   \\
    \texttt{hist\_ports}         & List      & Historical port sequence    & \texttt{["PortA", "PortB", "PortC"]} \\
    \texttt{future\_ports}         & List     & Target prediction sequence    & \texttt{["PortA", "PortB", "PortC"]} \\
    \bottomrule
  \end{tabularx}
\end{table}

\begin{figure}[h]
    \centering
    \includegraphics[width=1\linewidth]{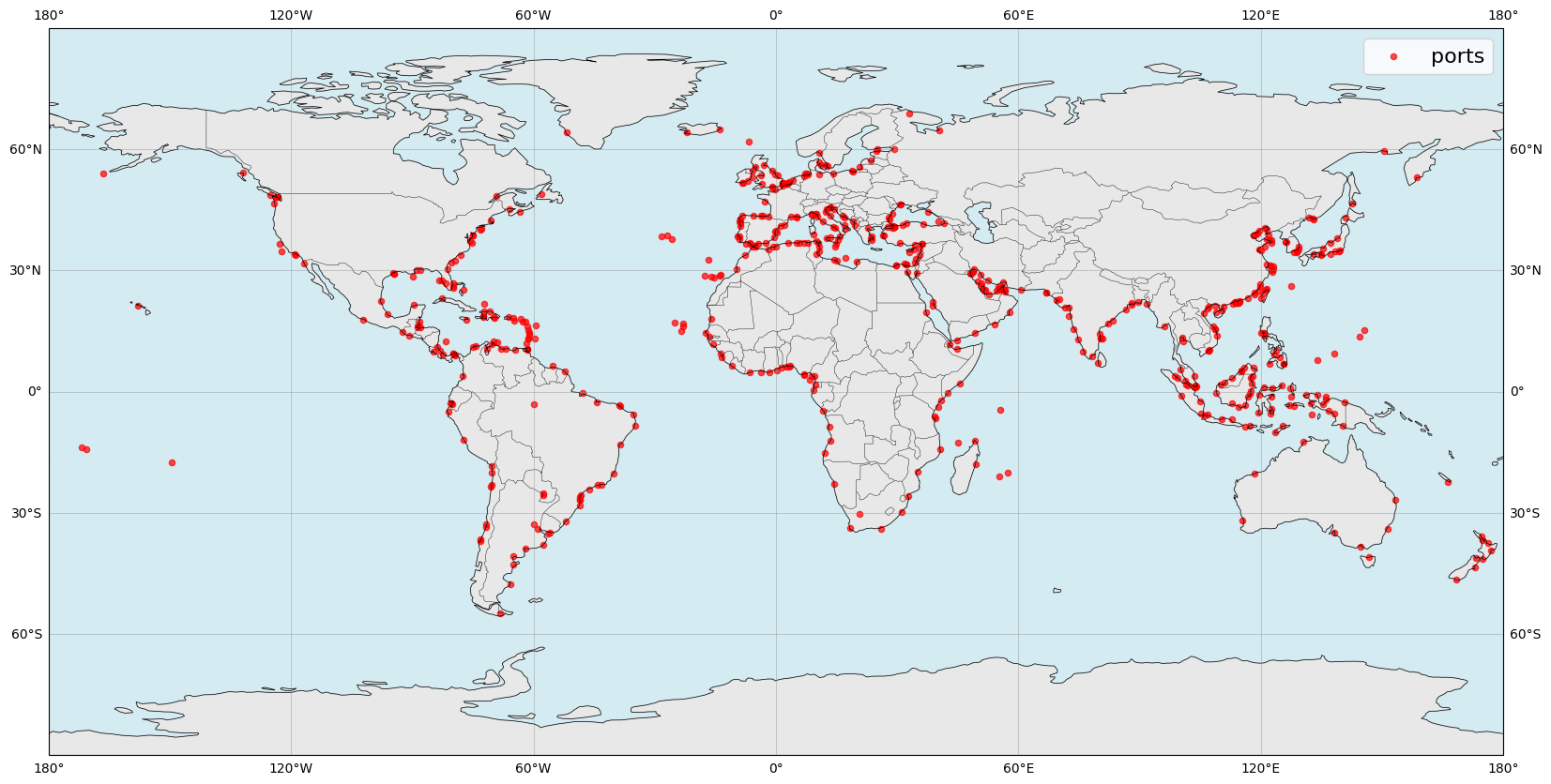}
    \caption{Global Ports Map}
    \label{fig:ports_map}
\end{figure}

Geographically, the dataset achieves comprehensive global coverage as visualized in Figure~\ref{fig:ports_map}. Network topology analysis identifies Singapore as the primary central hub, exhibiting the highest degree with direct connections to 69 distinct destinations.  The most heavily traversed edges characterize high-density short-haul corridors, such as Singapore to Port Klang (188 voyages) and Las Palmas to Santa Cruz (152 voyages). In terms of data integrity, the dataset maintains 100\% completeness across all essential feature fields. Crucially, regarding routing behaviors, approximately 16\% of the voyage segments exhibit cyclical port repetition within the prediction horizon ($H=3$). This characteristic provides a rigorous testbed for evaluating the model's capacity to capture recurrent operational patterns in liner shipping.

\subsection{Evaluation Metrics}\label{sec:metrics}
We evaluate the multi-step port prediction framework as a sequential classification task over the global port vocabulary $\mathcal{P}$ with a prediction horizon of $H=3$. 
For all metrics considered, we focus exclusively on the deterministic Top-1 prediction, denoted by:
\begin{equation*}
\hat{y}_{s,h} = \operatorname*{arg\,max}_{k \in \mathcal{R}_h(o_s)} \; p_{\theta}(k \mid \mathbf{z}^{(\mathrm{fuse})}_{s,h}, \hat{y}_{s,<h}),
\end{equation*}
which is the single destination candidate with the highest probability mass from the connectivity-constrained distribution to represent the model's most confident and physically feasible forecast.
In order to handle varying sequence lengths in the ground truth, we define $\mathcal{S}_h = \{s \mid y_{s,h} \neq \omega\}$ as the set of valid samples at step $h$ where the ground truth label is not the sentinel token $\omega$. 

To comprehensively assess the model performance, we first examine the step-wise accuracy at horizon $h$, denoted as $\text{Acc}_h$, which measures the proportion of valid samples where the predicted port matches the ground truth:
\begin{equation}\label{eq:step}
\mathrm{Acc}_h
=
\mathbb{E}_{s\sim\mathcal{S}_h}
\big[
\mathbf{1}\{\hat y_{s,h}=y_{s,h}\}
\big],
\end{equation}
where $\mathbb{I}(\cdot)$ is the indicator function. We specifically report $\text{Acc}_1$ to benchmark immediate predictive capability against single-step baselines. Furthermore, to provide a holistic assessment over the entire prediction horizon rather than focusing solely on the first step, we report the Average Accuracy ($\text{AvgAcc}$), defined as the unweighted mean of the stepwise accuracies across the $H$ steps:
\begin{equation}\label{eq:metric}
\text{AvgAcc} = \frac{1}{H} \sum_{h=1}^{H} \text{Acc}_h.
\end{equation}
Beyond pointwise correctness, we introduce Sequence Accuracy ($\text{SeqAcc}$) as a rigorous criterion for sequence-level comparison. Unlike point-wise metrics that allow partial matches, this metric treats the entire predicted route as a single evaluation unit, deeming a prediction successful if and only if the generated sequence perfectly aligns with the ground truth across all $H$ steps simultaneously. Letting $\mathcal{S}_{\text{seq}} = \bigcap_{h=1}^H \mathcal{S}_h$ be the set of samples with complete valid ground truth sequences, the metric is defined as:
\begin{equation}\label{eq:seq}
\mathrm{SeqAcc}
=
\mathbb{E}_{s\sim\mathcal{S}_{\mathrm{seq}}}
\left[
\prod_{h=1}^{H}
\mathbf{1}\{\hat y_{s,h}=y_{s,h}\}
\right].
\end{equation}
This multiplicative formulation makes $\text{SeqAcc}$ particularly sensitive to error accumulation, effectively reflecting the model's capability to maintain long-term routing logic.

For all accuracy metrics, we implement a systematic label-filtering procedure. Any (sample, step) pair whose target label is mapped to the sentinel index $\omega$ (unknown or out-of-vocabulary port) is excluded from both the numerator and denominator. Consequently, all reported accuracies are computed on the subset of positions with valid supervision. In addition, decoding is performed under the step-specific reachability masks, so only ports that are navigationally feasible are considered during prediction. All accuracy values are micro-averaged over the resulting set of valid (sample, step) pairs, yielding statistically stable estimates even when the proportion of valid labels differs across horizons.

In addition to accuracy, we monitor the label-smoothed cross-entropy loss introduced in Section~\ref{sec:learning-objective} as a scalar summary of predictive quality and calibration. The same loss function used for training—combining label smoothing, masking of sentinel labels, and connectivity-constrained softmax over $\mathcal{R}_h(o_s)$—is reported on the validation and test sets under autoregressive decoding. This provides a consistent objective measure that reflects not only classification accuracy but also the sharpness and calibration of the predicted distributions, while remaining faithful to the operational reachability constraints imposed by the maritime network.

\subsection{Experimental Settings}\label{sec:experimental-settings}
In this section, we will introduce the basic experiment setting, including training dataset splitting, parameters design and training strategies, and the baseline models for comparison.
\subsubsection{Chronological Data Splitting strategy.}\label{sec:split}
To simulate a realistic operational forecasting scenario, we reject random shuffling in favor of a strict chronological data splitting strategy. Instead, the dataset, covering the entirety of 2021, is partitioned into training, validation, and testing sets chronologically following a 70\%:15\%:15\% ratio. Consequently, the training set encompasses approximately the first 8.4 months of the year, capturing the foundational routing behaviors. The subsequent 15\% serves as the validation set for hyperparameter tuning, and the final 15\% is reserved strictly for testing to evaluate generalization capability on unseen future intervals. Regarding fleet composition, the splitting ensures that the vast majority of the 400 vessels are represented across all three partitions, ensuring model continuity. Exceptions are limited to a small minority of vessels whose available operational data is too sparse or confined to a single temporal window. 

To ensure that the model cannot access future information through the retrieval mechanism, we constructed the historical trajectory database exclusively using voyage segments from the training set. During the validation and testing phases, we do not incrementally update the database with test inputs so that the model must predict late-year voyages solely based on routing patterns learned from the early-to-mid-year history, thereby strictly testing its ability to generalize rather than memorize.

\subsubsection{Training setup and strategy.}
The training process was executed on a server equipped with an NVIDIA A100 GPU to accommodate the computational demands of the retrieval-enhanced encoder. Prior to training, continuous features (e.g., coordinates, speed) underwent robust Z-score normalization and were clipped to the range $[-10, 10]$ to mitigate the impact of sensor outliers. 

The model was trained end-to-end using an iterative paradigm over 50 epochs with a batch size of 64. We employed the Adam optimizer for parameter updates, initialized with a learning rate of $1 \times 10^{-4}$. To ensure training stability and generalization, we implemented a composite regularization strategy, including a weight decay of $1 \times 10^{-5}$ to penalize complex model weights, a Label Smoothing factor of $\varepsilon = 0.1$ to prevent over-confidence in the long-tail classification task, and a learning rate scheduler to monitor the validation accuracy. The learning rate is multiplied by a factor of 0.5 if the validation performance plateaus for 3 consecutive epochs, allowing for fine-grained convergence in later training stages.

To further reduce exposure bias in autoregressive decoding, we incorporated Scheduled Sampling~\citep{bengio2015scheduled} during training. At each prediction step, the decoder input from the previous step was selected as either the ground-truth token or the model-predicted token according to a time-varying sampling probability. The training process started with a high teacher-forcing ratio to ensure stable early optimization, and this ratio was gradually decreased over epochs so that the decoder increasingly relied on its own predictions. In this way, the model was progressively trained under conditions closer to inference, which improved robustness against error accumulation in multi-step port-sequence prediction. Specifically, the teacher-forcing ratio was linearly decayed from 1.0 to 0 over the training epochs.

\subsubsection{Baseline Models Design.}\label{sec:baseline_design}
We benchmark our framework against representative state-of-the-art models spanning two dominant paradigms. For feature-based classification, we employ RF, XGBoost, and CatBoost, selected for their robustness in handling structured AIS data. To capture the temporal dependencies inherent in multi-step forecasting, we encompass Seq2Seq architectures, specifically LSTM and Gated Recurrent Unit (GRU), serving as canonical deep learning benchmarks. This selection ensures a holistic comparison against both industry-standard tree ensembles and classical recurrent networks.

To rigorously benchmark these algorithms against our proposed framework, we employed distinct training strategies tailored to their architectural characteristics, while ensuring strictly fair comparison protocols. For non-sequential models (RF, XGBoost, CatBoost), we adapted them for the multi-step horizon ($H=3$) using a cascade strategy. This involves training $H$ independent classifiers, where the prediction from step $h-1$ is fed as an augmented feature to step $h$, thereby stimulating an auto-regressive process. We selected this approach over the independent strategy, which trains step-specific models in isolation, as preliminary experiments indicated a 1\% performance gain for the cascade method. For sequential models (LSTM, GRU), we utilized a standard Seq2Seq architecture trained end-to-end to autoregressively generate the future port sequence.


Finally, we structure the comparative analysis across three distinct dimensions of predictive fidelity to provide a comprehensive benchmarking of the proposed framework against these baselines. First, we evaluate the single-step accuracy ($\text{Acc}_1$), focusing strictly on the immediate next-port accuracy ($h=1$). This metric isolates the fundamental classification power of the models, decoupled from the effects of error propagation. Second, we assess multi-step accuracy by monitoring the performance decay across the full prediction horizon ($h=1, 2, 3$) and reporting the AvgAcc. This dimension rigorously tests the robustness of the cascade strategy (for tree ensembles) and the autoregressive mechanism (for sequence models) against compounding errors over time. Third, and most critically, we employ the SeqAcc to evaluate trajectory-level coherence. As the strictest criterion, this metric requires the perfect reconstruction of the entire future route, thereby penalizing any logical inconsistency or hallucination within the predicted sequence, ensuring that the model captures not just individual ports, but the holistic routing logic.
\section{Experiment Results and Analysis}\label{sec:experiment}
In this section, we present a comprehensive empirical evaluation of the proposed framework. We first report the comparative performance against the established baselines, validating the model's effectiveness and superiority in both single-step precision and multi-step trajectory consistency. Subsequently, we conduct rigorous ablation studies to dissect the impact of key architectural components and feature sets, thereby justifying the rationale behind our specific design choices. Finally, we synthesize the experimental findings to provide a holistic analysis of the model's behavior—particularly its robustness in long-tail scenarios—and outline potential avenues for future research.

\subsection{Baselines}\label{sec:baselines}
Following the evaluation protocols and baseline configurations outlined in Section~\ref{sec:baseline_design}, we first report the absolute performance of the proposed framework, followed by a systematic comparison against the established benchmarks. This comparative analysis evaluates the proposed framework across varying prediction horizons, benchmarking its performance against iterative tree ensembles and standard Seq2Seq recurrent architectures.

\begin{figure}[htbp]
    \centering
    \begin{subfigure}[b]{0.48\linewidth}
        \includegraphics[width=\linewidth]{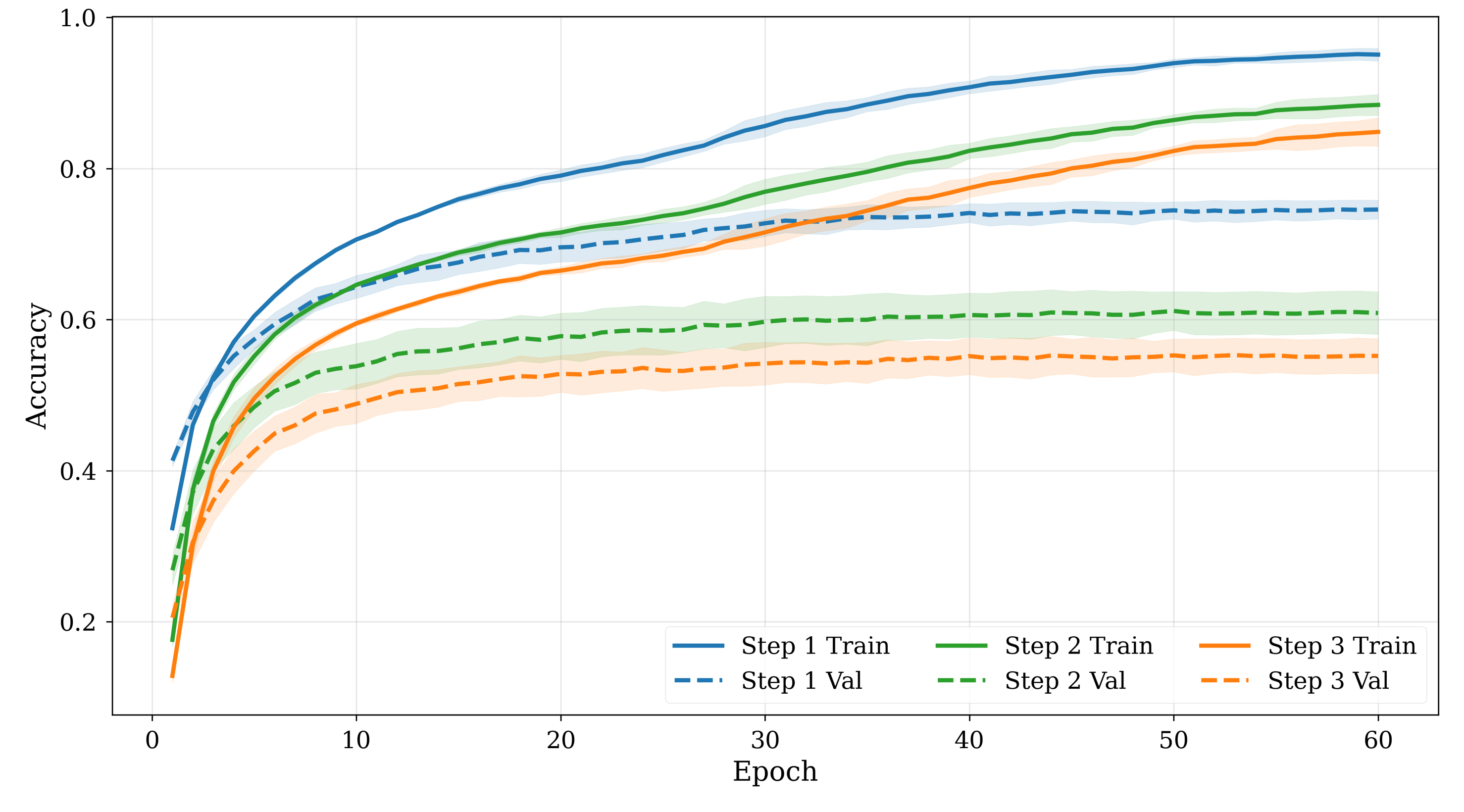}
        \caption{Step-wise Accuracy}
    \end{subfigure}
    \hfill
    \begin{subfigure}[b]{0.45\linewidth}
        \includegraphics[width=\linewidth]{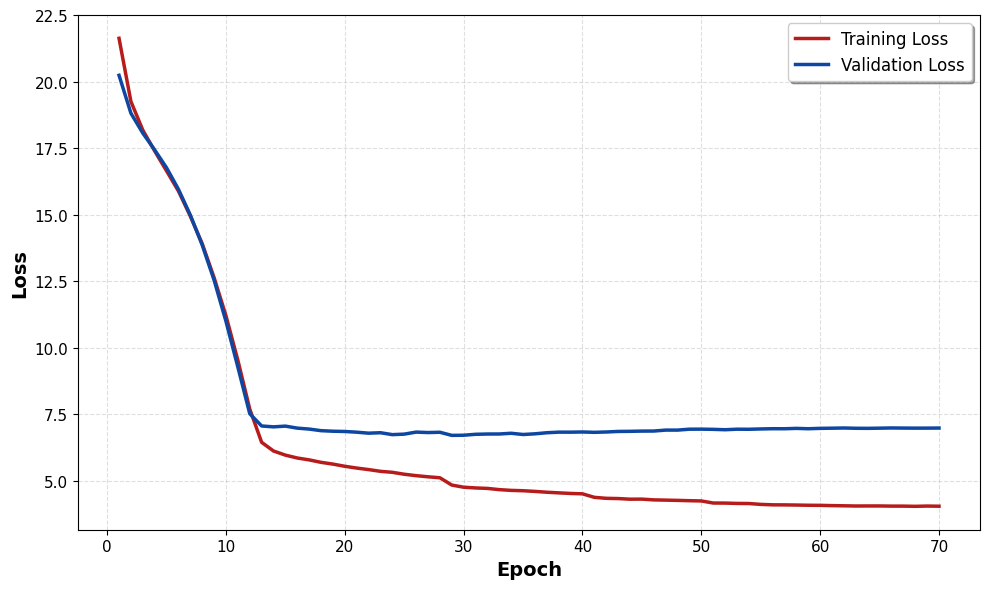}
        \caption{Training and Validation Loss}
    \end{subfigure}
    \caption{Training dynamics. The divergence between training (red) and validation (blue) curves highlights the distribution shift inherent in chronological data splitting.}
    \label{fig:training_curves}
\end{figure}

Overall, the proposed framework demonstrates robust predictive efficacy across the entire dataset. On the validation set, the model achieves high step-wise accuracies of $\text{Acc}_1=75.3\%$, $\text{Acc}_2=63.7\%$, and $\text{Acc}_3=57.1\%$, culminating in an  AvgAcc of $65.4\%$. Furthermore, it attains a SeqAcc of $46.01\%$, indicating a strong capability to capture coherent multi-step routing logic over the horizon $H=3$. These results are underpinned by the training dynamics shown in Figure~\ref{fig:training_curves}, which reveal a distinct two-phase convergence: a rapid initial feature adaptation phase followed by a stabilization period. The observed gap between training and validation performance reflects the intrinsic challenge of temporal generalization under strict chronological splitting, confirming that the model effectively learns foundational routing patterns despite the inevitable distribution shifts across disjoint time periods. 

When evaluated on the test set, the model achieves a $\text{Acc}_1=72.3\%$ and an $\text{AvgAcc}=61.4\%$, outperforming benchmarks by 12.6\% and 11.3\%  respectively. While this performance exhibits a moderate decline of approximately 3 to 4 percentage points compared to the validation metrics, we regard this generalization gap as acceptable given the strict chronological splitting strategy. Since the test set represents a more distant future relative to the training period, this slight degradation reflects the natural temporal drift inherent in maritime operations rather than a failure of model generalization.



Table~\ref{tab:baseline-results} reports the single-step predictive performance ($\text{Acc}_1$) on the full validation and test sets as defined in Eq.~\eqref{eq:step}. Among the baseline algorithms, CatBoost attains the highest accuracy (0.696 on validation and 0.644 on test), demonstrating its effectiveness in handling categorical port features. In contrast, RF and XGBoost, while comparable during validation, exhibit more significant degradation on the test set given the high-dimensional label space. Notably, the proposed framework significantly outperforms all baselines surpassing the strongest baseline by a substantial margin of +7.9\% on the test set. This superior single-step precision establishes a robust foundation for the subsequent error-sensitive multi-step autoregressive forecasting.


\begin{table}[h]
\centering
\small
\caption{Comparison of Single-step Accuracy ($\text{Acc}_1$) against baselines.}
\label{tab:baseline-results}
\begin{tabular}{lccccl}
\toprule
\multirow{2}{*}{Model} & \multicolumn{2}{c}{Validation} & \multicolumn{2}{c}{Test} & \multicolumn{1}{c}{Key settings} \\
\cmidrule(lr){2-3}\cmidrule(lr){4-5}
 & $\text{Acc}_1$ & Gap & $\text{Acc}_1$ & Gap & (abridged) \\
\midrule
\textbf{Proposed} & \textbf{0.753} & - & \textbf{0.723} & - & $m{=}3$, topology-masked \\
RF        & 0.646 & -10.7\% & 0.626 & -9.7\% & 200 trees, depth 10 \\
XGBoost   & 0.646 & -10.7\% & 0.582 & -14.1\% & 600 est., depth 8 \\
CatBoost  & 0.696 & -5.7\% & 0.644 & -7.9\% & 400 iters, depth 8 \\
\bottomrule
\end{tabular}
\end{table}

Table~\ref{tab:cascade-baselines} presents the performance trajectory across the full prediction horizon ($H=3$) and the average performance define in Eq.~\eqref{eq:metric}. This evaluation specifically assesses the models' resilience to error propagation, distinguishing our connectivity-constrained decoding from the unconstrained, self-feeding inference of the baselines. The proposed framework consistently outperforms all benchmark algorithms at every step. A critical observation is the rate of performance decay: while tree-based ensembles (e.g., RF and XGBoost) and recurrent models experience rapid degradation due to compounding errors in their unconstrained search spaces, our model maintains a significantly more robust performance profile.


\begin{table}[h]
\centering
\small
\caption{Comparison of Multi-step Accuracy against baselines.}
\label{tab:cascade-baselines}
\begin{tabular}{lcccccccc}
\toprule
\multirow{2}{*}{Model} & \multicolumn{4}{c}{Validation} & \multicolumn{4}{c}{Test} \\
\cmidrule(lr){2-5}\cmidrule(lr){6-9}
 & $\text{Acc}_1$ & $\text{Acc}_2$ & $\text{Acc}_3$ & $\text{AvgAcc}$ & $\text{Acc}_1$ & $\text{Acc}_2$ & $\text{Acc}_3$ & $\text{AvgAcc}$ \\
\midrule
\textbf{Proposed} & \textbf{0.753} & \textbf{0.637} & \textbf{0.571} & \textbf{0.654} & \textbf{0.723} & \textbf{0.595} & \textbf{0.525} & \textbf{0.614} \\
RF        & 0.674 & 0.498 & 0.433 & 0.535 & 0.666 & 0.467 & 0.415 & 0.516 \\
XGBoost   & 0.654 & 0.530 & 0.481 & 0.555 & 0.588 & 0.477 & 0.417 & 0.494 \\
CatBoost  & 0.696 & 0.590 & 0.527 & 0.604 & 0.654 & 0.533 & 0.459 & 0.549 \\
LSTM      & 0.578 & 0.502 & 0.469 & 0.516 & 0.536 & 0.461 & 0.422 & 0.473 \\
GRU       & 0.583 & 0.501 & 0.467 & 0.517 & 0.541 & 0.459 & 0.418 & 0.473 \\
\bottomrule
\end{tabular}
\end{table}

Table~\ref{tab:main_results} summarizes the performance using the most rigorous criterion, $\text{SeqAcc}$ as in Eq~\eqref{eq:seq}, which requires the perfect reconstruction of the entire trajectory over the horizon $H=3$. Among the baselines, CatBoost exhibits the most robust generalization. Notably, generic recurrent architectures underperform the strongest tree baseline, indicating a struggle to capture the sparse, long-tail distribution of global shipping routes without explicit structural priors. In contrast, the proposed framework consistently outperforms all benchmarks.

\begin{table}[htbp]
  \centering
  \caption{Comparison of Sequence Accuracy (\text{SeqAcc}) against baselines.}
  \label{tab:main_results}
  \begin{tabular}{llcc}
    \toprule
    \textbf{Model} & \textbf{Strategy} & \textbf{Val \text{SeqAcc}} & \textbf{Test \text{SeqAcc}} \\
    \midrule
    \textbf{Proposed} & - & \textbf{0.4601} & \textbf{0.4139} \\
    \midrule
    RF & Cascade & 0.2965 & 0.2730 \\
    XGBoost       & Cascade & 0.3641 & 0.2925 \\
    CatBoost      & Cascade & 0.4537 & 0.3919 \\
    \midrule
    LSTM          & Seq2Seq & 0.3114 & 0.2936 \\
    GRU           & Seq2Seq & 0.3302 & 0.3042 \\
    \bottomrule
  \end{tabular}
\end{table}

\subsection{Stratified Analysis}
To demonstrate that our proposed architecture exhibits resilience against data long-tail and sparsity characteristics, we conducted a hierarchical analysis based on the baseline design described earlier.
In fact, global shipping networks exhibit a severe long-tail distribution, where a small fraction of ``head'' ports dominates traffic, while the majority of ``tail'' ports are visited infrequently. As shown in our dataset, the top 20\% of ports account for 66.7\% of total calls, whereas the bottom 30\% account for only 3.0\%. Standard data-driven models often struggle in these sparse regions due to label bias. To evaluate our model's robustness, we stratified the test results by target port popularity, comparing our approach against benchmarks, the results are as follows in Table ~\ref{tab:long_tail_analysis_all}.

\begin{table*}[h]
  \centering
  \caption{Performance stratification by port popularity across different models.}
  \label{tab:long_tail_analysis_all}
  \resizebox{\textwidth}{!}{%
    \begin{tabular}{llcccccc}
    \toprule
    \textbf{Stratum} & \textbf{Metric} & \textbf{Proposed} & \textbf{CatBoost} & \textbf{XGBoost} & \textbf{RF} & \textbf{LSTM} & \textbf{GRU} \\
    \midrule
    \textbf{Head (Top 20\%)} & $\text{Acc}_1$ & \textbf{73.73\%} & 67.68\% & 63.95\% & 58.07\% & 48.98\% & 48.81\% \\
    \textit{Samples: 2,893} & $\text{Acc}_2$ & \textbf{58.04\%} & 54.99\% & 50.49\% & 47.68\% & 41.75\% & 43.24\% \\
    \textit{Freq: 66.7\%} & $\text{Acc}_3$ & \textbf{50.28\%} & 49.03\% & 43.01\% & 42.24\% & 40.01\% & 39.14\% \\
    \midrule
    \textbf{Body (Mid 50\%)} & $\text{Acc}_1$ & \textbf{69.48\%} & 61.35\% & 48.41\% & 36.95\% & 42.14\% & 42.14\% \\
    \textit{Samples: 1,291} & $\text{Acc}_2$ & \textbf{55.67\%} & 50.61\% & 41.72\% & 29.75\% & 39.49\% & 39.42\% \\
    \textit{Freq: 30.3\%} & $\text{Acc}_3$ & \textbf{49.16\%} & 46.18\% & 38.23\% & 27.75\% & 33.26\% & 35.32\% \\
    \midrule
    \textbf{Tail (Bottom 30\%)} & $\text{Acc}_1$ & \textbf{40.12\%} & 32.72\% & 22.22\% & 8.64\% & 14.81\% & 18.52\% \\
    \textit{Samples: 162} & $\text{Acc}_2$ & \textbf{31.75\%} & 29.38\% & 24.05\% & 5.06\% & 8.23\% & 20.25\% \\
    \textit{Freq: 3.0\%} & $\text{Acc}_3$ & \textbf{20.00\%} & \textbf{20.00\%} & 16.36\% & 3.03\% & 9.09\% & 13.94\% \\
    \bottomrule
    \end{tabular}%
  }
\end{table*}

While CatBoost performs commendably on head ports ($\text{Acc}_1=67.68\%$) due to its optimized handling of categorical features, it struggles to generalize to the ``body'' stratum, dropping by 6.33\% to 61.35\%. In contrast, our proposed model exhibits a significantly flatter degradation curve. It maintains a high accuracy of 69.48\% in the body stratum (only a 4.25\% drop from head), outperforming CatBoost by 8.13 percentage points. This result suggests that the retrieval-enhanced encoder effectively transfers navigation patterns from dominant routes to medium-frequency ones, mitigating the overfitting often observed in pure tree-based ensembles. By retrieving similar routes from historical databases $\mathcal{H}$, the model can identify reference data that scarce but holds greater significance. It enhances attention allocation to these instances, ultimately improving prediction outcomes.

The efficacy of the connectivity-constrained encoding mechanism is most evident in the sparse ``tail'' stratum. At step 1, our model achieves an accuracy of 40.12\%, surpassing CatBoost (32.72\%) by +7.4\% and nearly doubling the performance of XGBoost (22.22\%). Although CatBoost is a robust statistical learner, it lacks explicit knowledge of physical reachability. By strictly constraining the search space to navigationally feasible neighbors, our model effectively ``prunes'' implausible predictions for rare ports. This establishes a robust performance floor even when historical data is scarce, ensuring operational feasibility in long-tail scenarios where purely data-driven baselines tend to fail.


In summary, the comparative analysis across single-step, multi-step cascade, and sequence-level evaluations reveals distinct behavioral differences between the proposed framework and existing baselines.  First, in terms of single-step precision, the model effectively synergizes immediate trajectory perception with historical attention mechanisms. By attending to relevant past voyages, the encoder enhances the representation of the current vessel state, thereby significantly boosting the first step accuracy ($\text{Acc}_1$). 
Second, for multi-step precision, the architecture leverages the Transformer's global receptive field to synthesize long-term temporal dependencies. The autoregressive generation mechanism transforms historical sequential patterns into effective future references, enabling the coherent inference of the port sequence $y_{t+1:t+H}$ by modeling the conditional probability distribution $P(y_{t+h} \mid y_{<t+h}, \mathbf{X})$.
Finally, regarding robustness to long-tail sparsity, the framework addresses the label bias inherent in maritime data through a dual mechanism: the retrieval-enhanced encoder reinforces feature extraction for sparse historical samples, while the connectivity-constrained decoding explicitly prunes the search space. By confining predictions to the valid reachability set $\mathcal{R}_h$, the model successfully avoids infeasible solutions, establishing a robust performance floor even for rare ports with limited training instances.


\subsection{Ablation Studies}\label{sec:ablation}
We conduct comprehensive ablation experiments to systematically evaluate the contribution of different model components, with particular focus on the historical retrieval pathway and individual feature modalities. Our analysis employs two complementary experimental designs: architectural component ablation and feature importance analysis.


\subsubsection{Architectural Component Ablation.}
To strictly quantify the incremental value of the retrieval-enhanced pathway, we evaluated the proposed architecture against four degraded variants defined by specific coding constraints. 

The first and second variants, denoted as the zero and random models, serve as lower bounds where the retrieval channel is either masked entirely or replaced with Gaussian noise. The third variant, referred to as the static model, encodes the input historical sequence directly but is restricted from querying the external database. This setup forces the model to rely solely on the internal recurrence of the past trajectory. The fourth variant, termed the no-similarity model, performs database queries but aggregates all retrieved future scenarios using uniform weights, thereby disabling the ability of the mechanism to prioritize specific historical routes.

The predicted view was fixed at H = 3 for all experiments, and $R_h(o_s)$ was applied as a reachability mask on the decoder. To maximize statistical sensitivity and quantify the intrinsic representational capacity of the retrieval mechanism independently of data splitting noise, the complete dataset was used as the experimental data.

The results (Table \ref{tab:ablation-architectural}) reveal a distinct performance hierarchy that aligns with the information-theoretic capacity of each variant. The static variant performs nearly identically to the  zero and  random models, with all three stalling at an AvgAcc of approximately 0.55. This stagnation indicates that the static encoder provides highly redundant information because the primary trajectory encoder already processes the spatiotemporal path of the vessel. The inability of the static model to outperform the zero baseline confirms that simply encoding the past history of the vessel is insufficient for multi-step forecasting. The model explicitly requires external knowledge regarding the future movements of similar vessels, which is unavailable through self-recurrence alone.

The transition from the static variant to the no-similarity scheme yields a substantial gain of approximately 20\% points, isolating the fundamental value of information retrieval. This significant improvement confirms that accessing precedents provides a far stronger predictive signal than merely encoding the vessel's current trajectories. Furthermore, the full model outperforms the no-similarity variant by an additional 7\%. This increment validates the efficacy of the similarity matching mechanism, which prioritizes semantically relevant scenarios. By increasing the weight of historical information with higher similarity, the attention mechanism effectively suppresses noise and mitigates error accumulation in autoregressive generation.

\begin{table}[h]
\centering
\caption{Architectural component ablation on the historical retrieval module.}
\label{tab:ablation-architectural}
\begin{tabular}{lcccc}
\toprule
\multirow{2}{*}{\textbf{Setting}} & \multirow{2}{*}{\textbf{AvgAcc}} & \multicolumn{3}{c}{\textbf{Step-wise Acc}} \\
\cmidrule(lr){3-5}
 & & $\textbf{Acc}_1$ & $\textbf{Acc}_2$ & $\textbf{Acc}_3$ \\
\midrule
\textbf{full} (Proposed) & \textbf{0.8290} & \textbf{0.8946} & \textbf{0.8189} & \textbf{0.7735} \\
zero & 0.5572 & 0.7569 & 0.5002 & 0.4146 \\
random & 0.5595 & 0.7544 & 0.5026 & 0.4216 \\
static & 0.5585 & 0.7546 & 0.5026 & 0.4183 \\
no\_similarity & 0.7593 & 0.8495 & 0.7444 & 0.6841 \\
\bottomrule
\end{tabular}
\end{table}

\subsubsection{Feature Importance Analysis.}

Complementing the architectural ablation, this phase isolates the contributions of heterogeneous data sources through systematic feature randomization on the test set. Specifically, we target the distinct feature groups that constitute the input to the overall encoding pipeline in Eq.~\eqref{eq:overall_pipeline}. By selectively corrupting these features we quantify their individual marginal contributions to the inference process. The results, summarized in Table~\ref{tab:ablation-features}, demonstrate a distinct mechanism of temporal feature specialization.

\begin{table}[h]
\centering
\small
\caption{Feature modality ablation through systematic randomization on test set.}
\label{tab:ablation-features}
\begin{tabular}{lcccc}
\toprule
\multirow{2}{*}{\textbf{Component}} & \multirow{2}{*}{\textbf{AvgAcc}} & \multicolumn{3}{c}{\textbf{Step-wise Acc}} \\
\cmidrule(lr){3-5}
 & & $\textbf{Acc}_1$ & $\textbf{Acc}_2$ & $\textbf{Acc}_3$ \\
\midrule
\textbf{full} (Proposed) & \textbf{0.5948} & \textbf{0.7225} & \textbf{0.5630} & \textbf{0.4989} \\
\midrule
fusion ($\mathbf{z}_{\mathrm{fused}}$)& 0.0961 & 0.1451 & 0.0795 & 0.0636 \\
\midrule
feature\_hist ($\mathbf{z}_{\mathrm{hist}}$)& 0.4363 & 0.6340 ($\downarrow$12.3\%) & 0.3707 ($\downarrow$34.2\%) & 0.3042 (\textbf{$\downarrow$39.0\%}) \\
feature\_kinematic ($\bm{X}^{(\mathrm{kin})}_s$) & 0.4095 & 0.5008 ($\downarrow$30.7\%) & 0.3828 (\textbf{$\downarrow$32.0\%}) & 0.3450 ($\downarrow$30.8\%) \\
feature\_dynamic ($\bm{X}^{(\mathrm{dyn})}_s$)& 0.3957 & 0.4704 (\textbf{$\downarrow$34.9\%}) & 0.3753 ($\downarrow$33.3\%) & 0.3412 ($\downarrow$31.6\%) \\
feature\_static ($\bm{X}^{(\mathrm{stat})}_s$)& 0.3941 & 0.4974 ($\downarrow$31.2\%) & 0.3685 ($\downarrow$34.5\%) & 0.3165 ($\downarrow$36.6\%) \\
\bottomrule
\end{tabular}
\end{table}


First, randomizing the fusion layer yields near-random performance with an AvgAcc of approximately 0.09. Given that the fusion module acts as the sole information bottleneck channeling inputs to the decoder, this performance collapse confirms that the model actively conditions predictions on the input context $P(Y|X)$ rather than simply memorizing the marginal distribution $P(Y)$ of the training labels.

Second, the ablation study highlights the critical function of static attributes ($\bm{X}^{(\mathrm{stat})}_s$) as global conditioning constraints. It is notable that randomizing time-independent features resulted in the lowest AvgAcc (0.3941) among all single modality ablations, comparable to the loss of dynamic kinematic signals. This severe degradation across all prediction steps suggests that the model relies heavily on vessel profiling to narrow the feasible solution space. For instance, two containers with different TEU capacity may share identical coordinates and speeds in open water, yet their destination probabilities are structurally distinct due to port compatibility and service loops. The performance drop confirms that the architecture effectively encodes these semantic constraints, using static embeddings as a prior distribution that guides the entire trajectory generation process.

Finally, the contribution of distinct modalities diverges as the prediction horizon extends. 
For short-term prediction ($\text{Acc}_1$), kinematic features ($\bm{X}^{(\mathrm{kin})}_s$) are dominant. Randomizing trajectory features reduces accuracy to 0.5008, which is significantly lower than the 0.6340 observed when randomizing the historical encoder. This confirms that the immediate physical state drives short-term trajectory generation. Conversely, for long-term prediction ($\text{Acc}_3$), historical context ($\mathbf{z}_{\mathrm{hist}}$) becomes indispensable. Randomizing the historical encoder causes accuracy to drop to 0.3042, a value noticeably lower than the 0.3450 resulting from kinematic randomization. 
Consequently, this crossover phenomenon indicates that while kinematic features serve as the primary driver for short-term predictions, its predictive contribution diminishes over time. In contrast, historical features maintains its validity as the horizon extends, thereby becoming the dominant signal for long-term forecasting.


Collectively, these ablation studies validate the proposed framework as a synergistic system characterized by structural conditioning and temporal specialization. The analysis confirms that the model effectively integrates heterogeneous inputs according to their distinct predictive roles. Static attributes enforce global feasibility constraints, while dynamic signals are utilized based on the forecast horizon. Furthermore, kinematic features drive immediate, physics-constrained driving, and the historical context serves as the strategic guide for long-term routing.
Moreover, the superiority of the full model over the static and no-similarity variants proves that such long-horizon guidance requires active, similarity-based retrieval rather than passive historical encoding. These findings emphasize the importance of a hybrid mechanism that utilizes the precision of real-time kinematics and the foresight of retrieved navigational precedents in order to achieve robust multi-step prediction.

\subsection{Case Studies of Route Patterns}\label{sec:case-study}

\begin{figure*}[!htbp]
    \centering
    
    \begin{subfigure}[b]{0.48\textwidth}
        \centering
        \includegraphics[width=\linewidth]{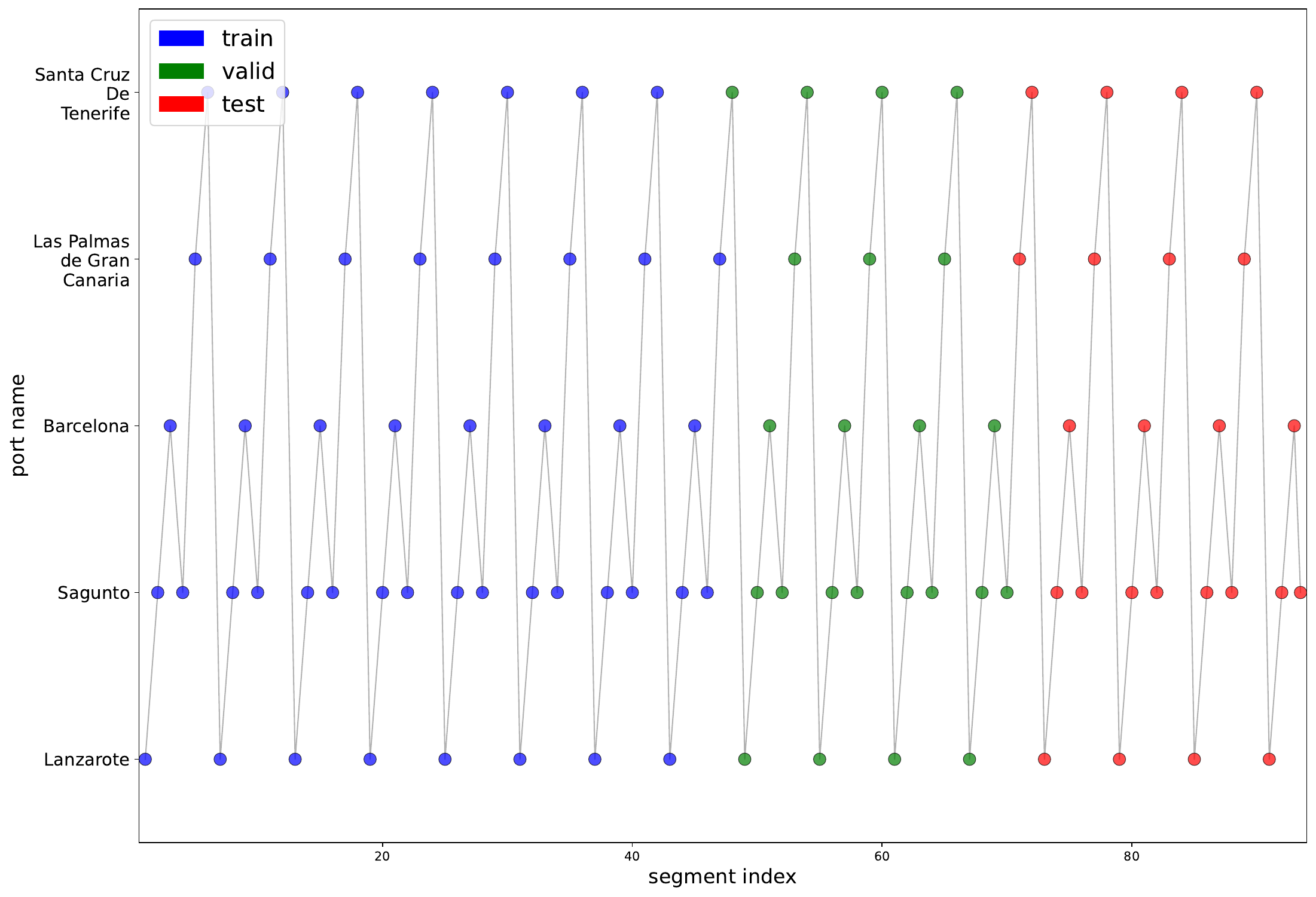}
        \caption{Cyclic Route 9144718}
        \label{fig:9144718}
    \end{subfigure}
    \hfill
    \begin{subfigure}[b]{0.48\textwidth}
        \centering
        \includegraphics[width=\linewidth]{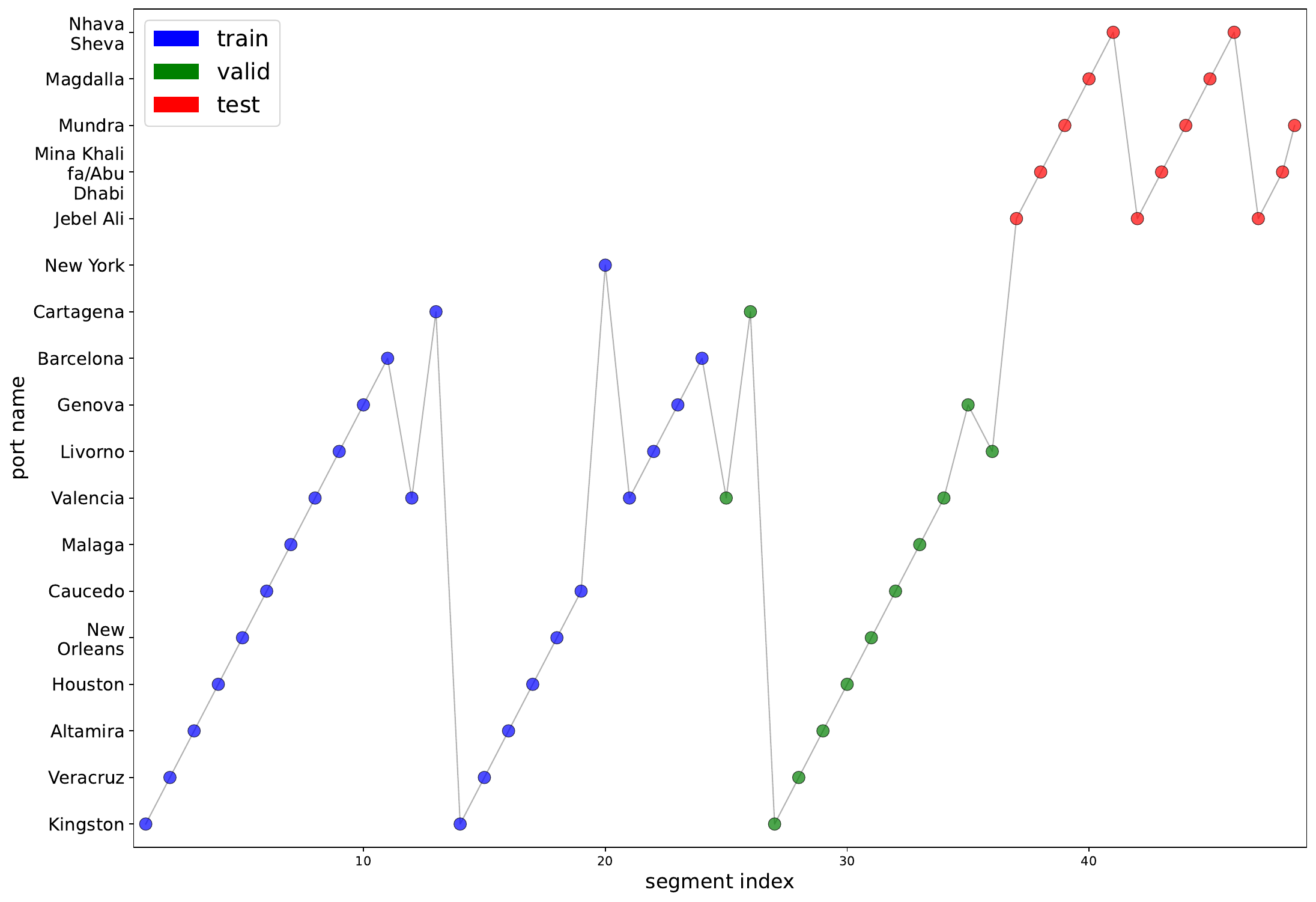}
        \caption{Time-evolving 9450301}
        \label{fig:9450301}
    \end{subfigure}

    \vspace{1em} 

    \begin{subfigure}[b]{0.48\textwidth}
        \centering
        \includegraphics[width=\linewidth]{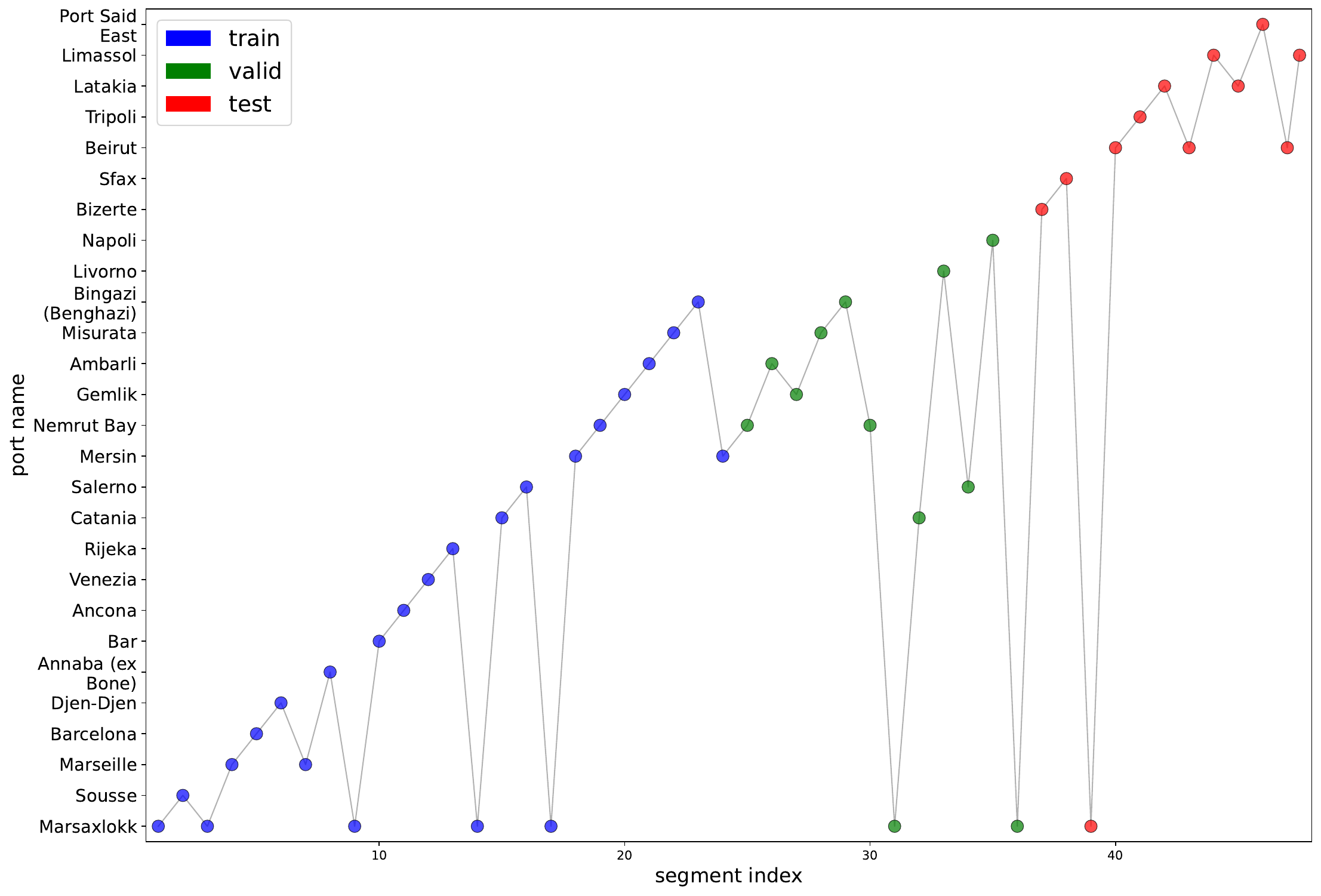}
        \caption{Irregular Route 9160920}
        \label{fig:9160920}
    \end{subfigure}
    \hfill
    \begin{subfigure}[b]{0.48\textwidth}
        \centering
        \includegraphics[width=\linewidth]{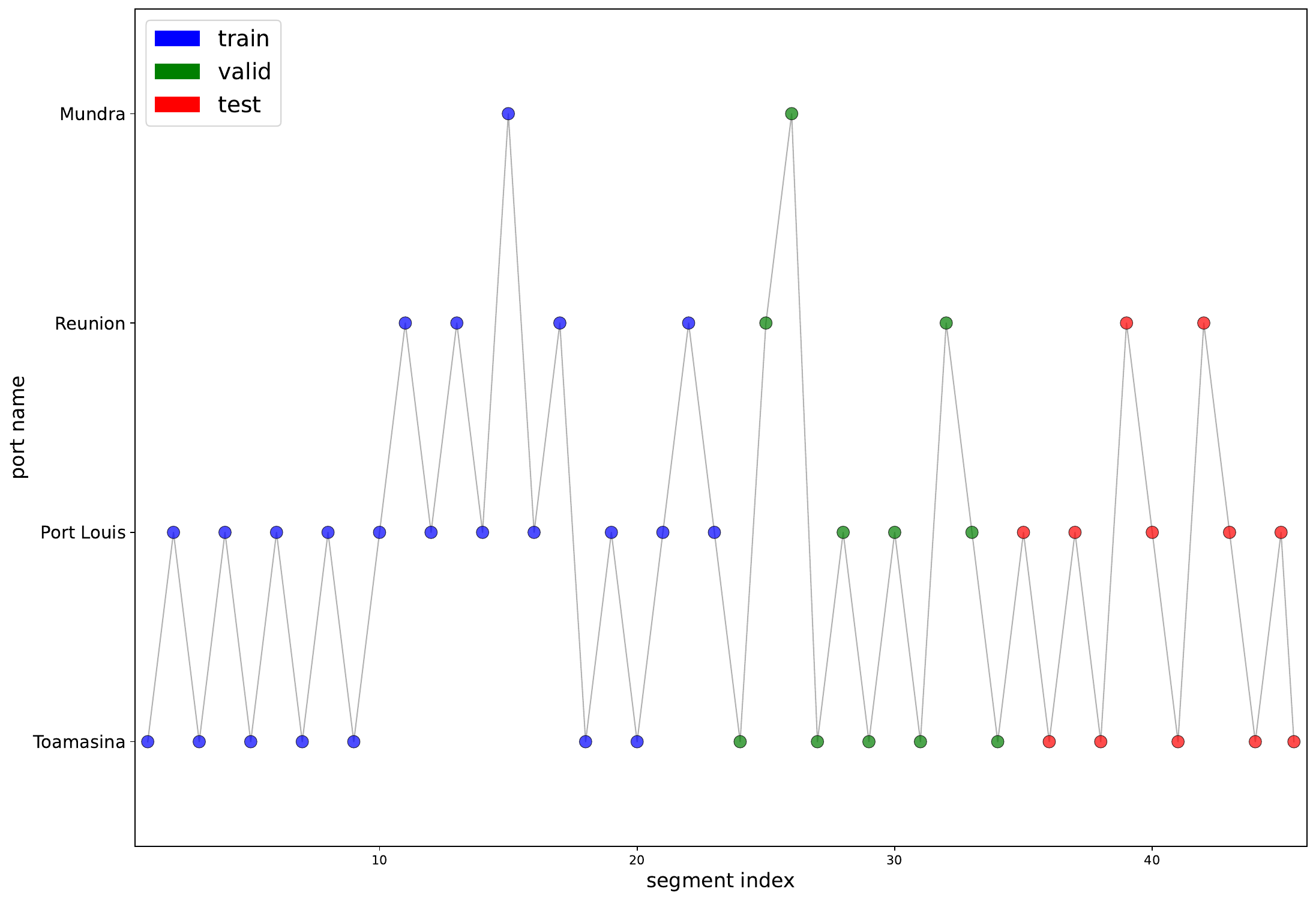}
        \caption{Partially Regular 8201624}
        \label{fig:8201624}
    \end{subfigure}
    
    \caption{Visualization of representative vessel route patterns. (a) Highly regular cyclic route; (b) Temporally evolving route; (c) Highly irregular route; (d) Partially regular route.}
    \label{fig:route_patterns}
\end{figure*}

To better understand how the proposed model behaves under different operational regimes, we conduct a set of vessel-level case studies. 
The routes are categorized into four types: (i) Cyclic routes: highly regular cyclic routes with near-perfect predictive performance, (ii) Time-evolving routes: temporally evolving routes whose behavior changes markedly with time, (iii) Irregular routes: highly irregular routes that are effectively unpredictable, and (iv) Other routes: partially regular routes that lie between the above extremes.
Table~\ref{tab:case-types} summarizes the $\text{SeqAcc}$ accuracies on validation and test sets and some representative vessels are shown in Figure~\ref{fig:route_patterns}.

\begin{table}[t]
\centering
\caption{Representative vessels and route patterns}
\label{tab:case-types}
\begin{tabular}{llllcc}
\toprule
\multirow{2.5}{*}{\textbf{Category}} & \multirow{2.5}{*}{\textbf{IMO}} & \multirow{2.5}{*}{\textbf{Route Summary}} & \multicolumn{2}{c}{\textbf{SeqAcc (\%)}} \\
\cmidrule(l){4-5} 
 & & & \textbf{Val} & \textbf{Test} \\
\midrule
Cyclic route & 9065431  &
Simple point-to-point shuttle
& 100 & 100 \\
Cyclic route & 9144718  &
Regional multi-port loop
& 100 & 100 \\
Time-evolving & 9450301  &
Stable (Val) $\to$ Shift (Test)
& 83.3 & 0.0 \\
Time-evolving & 9226827  &
Stable (Train) $\to$ Shift (Val)
& 0.0 & 0.0 \\
Irregular & 9160920  &
Irregular long-haul routing
& 0.0 & 0.0 \\
Partially regular & 8201624 &
Recurring loop with branching
& 22.2 & 12.5 \\
Partially regular & 9231248  &
Long-range inter-regional shuttle
& 45.5 & 37.5 \\
\bottomrule
\end{tabular}
\end{table}

\begin{figure}[h]
    \centering
    \includegraphics[width=1\linewidth]{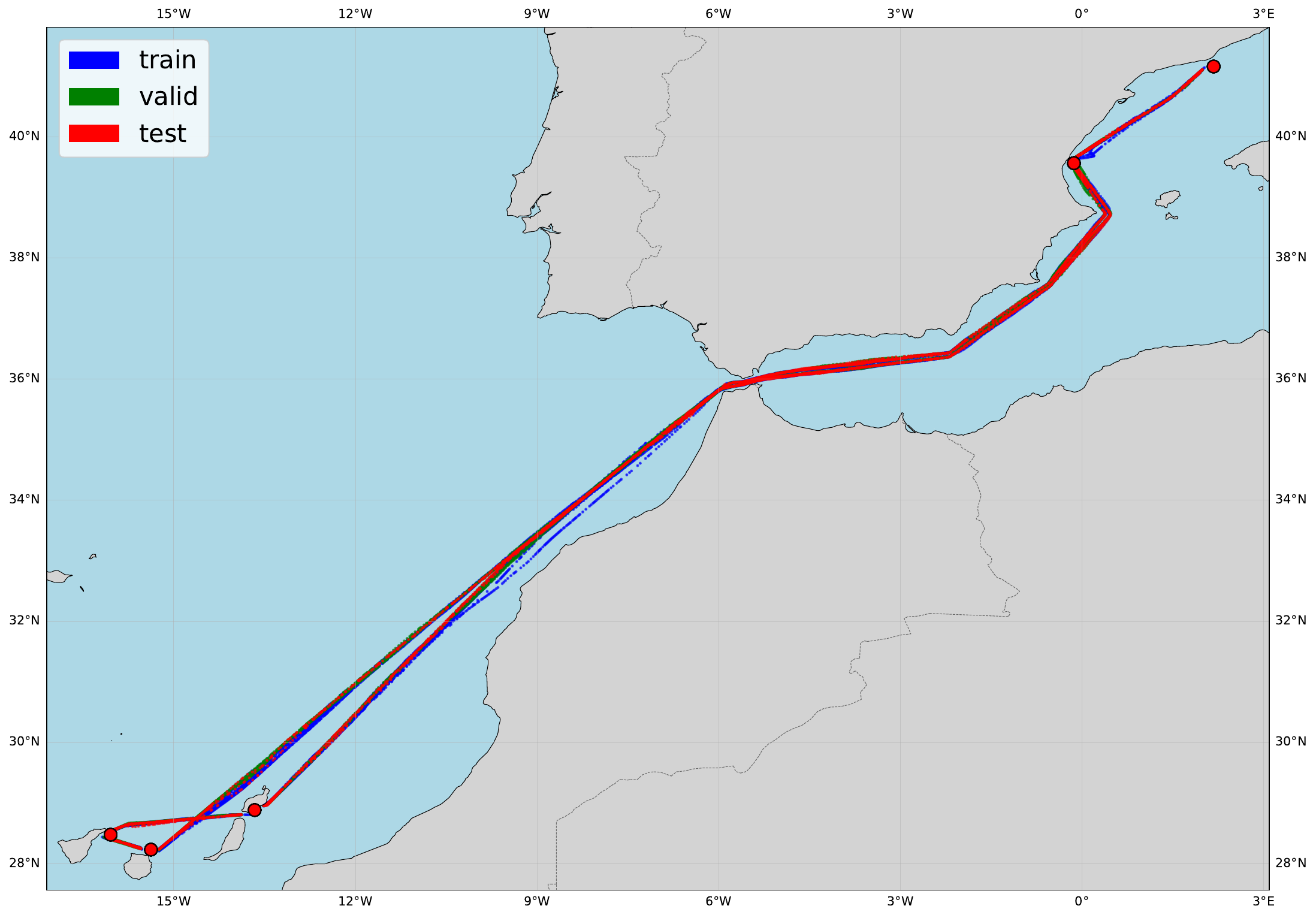}
    \caption{The trajectory of vessel 9144718}
    \label{fig:traj_9144718}
\end{figure}

For cyclic routes, a canonical short-range cyclic example is IMO~9065431, which operates an extremely stable two-port shuttle between Jakarta and Panjang (hereinafter $J \& P$). 
Across 57 prediction instances (23 validation, 34 test) the model achieves 100\% $\text{SeqAcc}$ and perfectly recovers the underlying cycle. This is because the observed movement pattern can be summarized as closed loops: $J \rightarrow P \rightarrow J \rightarrow P \rightarrow \cdots$
Once the model sees a short history such as
$J$~$\rightarrow$~$P$~$\rightarrow$~$J$, 
it reliably predicts the next three ports as
$P$~$\rightarrow$~$J$~$\rightarrow$~$P$,
matching the ground truth exactly for every occurrence.
A more complex cyclic behaviour is exhibited by IMO~9144718 (Figure~\ref{fig:9144718}), which operates a fixed multi-port loop as show in Figure~\ref{fig:traj_9144718}: $\text{Sagunto} \rightarrow \text{Barcelona} \rightarrow \text{Sagunto} \rightarrow \text{Las Palmas de Gran Canaria}
\rightarrow \text{Santa Cruz de Tenerife} \rightarrow \text{Lanzarote} \rightarrow \text{Sagunto} \rightarrow \cdots.$
Variations of this loop appear in both directions, with the vessel sometimes entering the cycle from the Canary side and returning to the Barcelona--Sagunto pair.
Despite the richer structure (six distinct modes within a fixed port set), the model again attains 100\% $\text{SeqAcc}$. The cyclic routes indicate that once the port set and loop structure are fixed, the multi-modal historical encoder and connectivity graph can capture the recurrent patterns robustly.


\begin{figure}[h]
    \centering
    \includegraphics[width=1\linewidth]{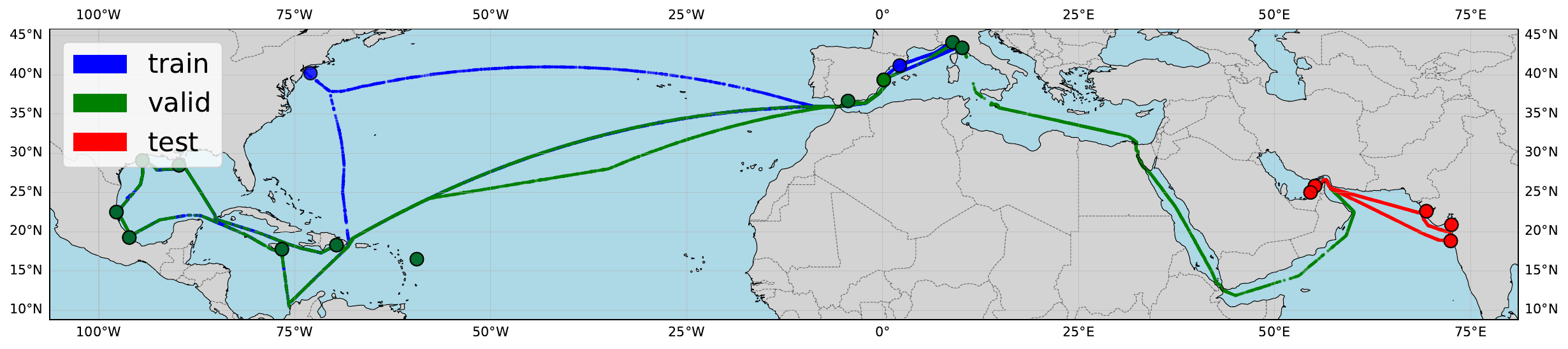}
    \caption{The trajectory of vessel 9450301}
    \label{fig:traj_9450301}
\end{figure}

The second category captures vessels whose routing behaviour evolves substantially between the validation and test periods. 
At the aggregate level, several ships exhibit large validation--test gaps in $\text{SeqAcc}$. 
IMO~9450301 (Figure~\ref{fig:9450301}) is a prototypical example where the route appears regular in the earlier period but changes markedly later. The trajectory details are illustrated in Figure~\ref{fig:traj_9450301}.
In the validation split, the vessel behaves almost deterministically from the model's perspective, with $\text{SeqAcc}$ reaching 83.3\%, and most errors limited to minor deviations in later steps. 
However, in the test split, the $\text{SeqAcc}$ drops to 0\%, and even per-step accuracies become very low: the model is occasionally correct at the first step but systematically fails to follow the new routing logic thereafter. 
Another complementary example is IMO~9226827. While the vessel exhibits structured routing patterns during training, it suffers from 0\% $\text{SeqAcc}$ in both validation and test phases. The model frequently predicts functionally similar ports within the same coastal corridor rather than the exact ground truth. This indicates that in competitive regions, the specific service loop is highly dynamic, preventing the fixed historical logic learned during training from generalizing to subsequent time windows. Overall, the temporally evolving routes highlight a limitation of static training. The model can fit a stable pattern well in one time window, but generalization declines once the operator adjusts schedules.

\begin{figure}[h]
    \centering
    \includegraphics[width=1\linewidth]{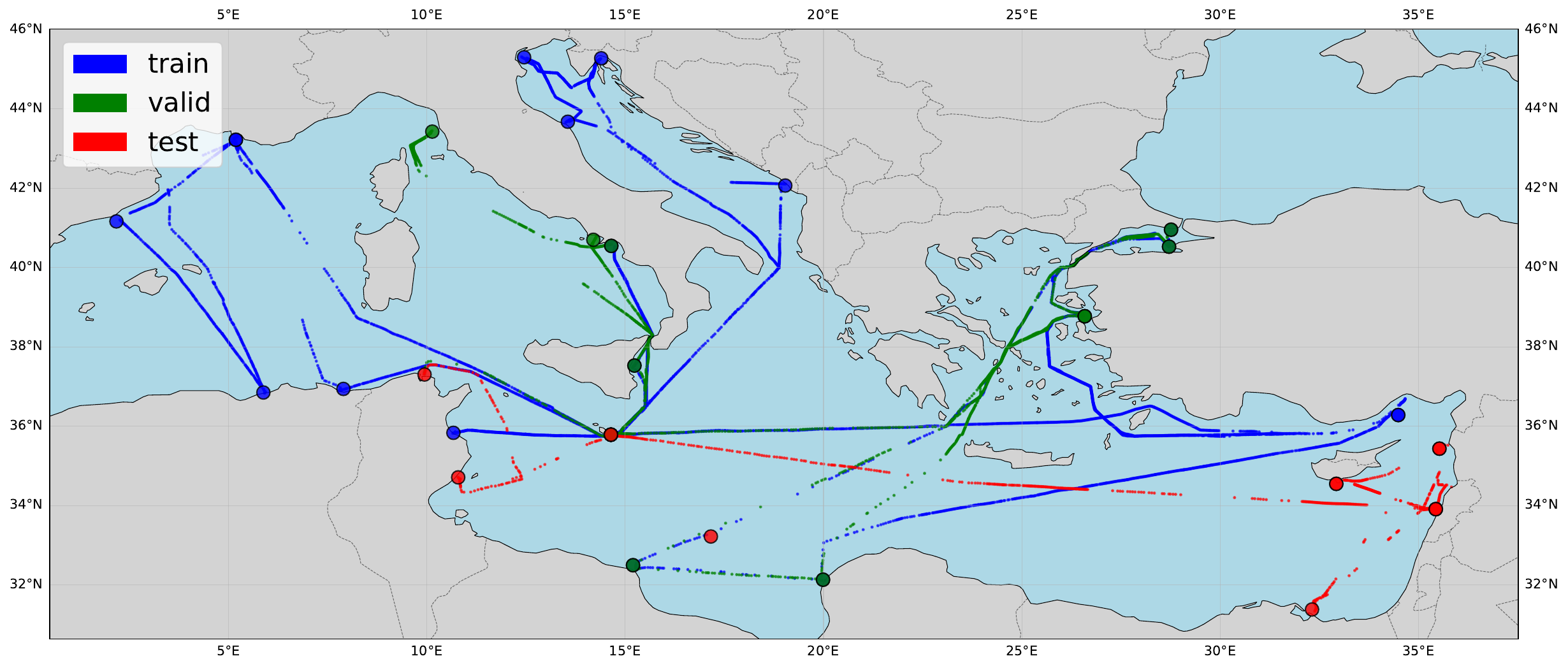}
    \caption{The trajectory of vessel 9160920}
    \label{fig:traj_9160920}
\end{figure}

Furthermore, for vessels in the irregular category, their routes are entirely unstructured, exhibiting no reusable pattern over the entire horizon. A representative example is IMO~9160920 (Figure~\ref{fig:9160920}), which operates highly irregular long-distance routes across the Mediterranean. The vessel performs long-distance movements linking ports such as Bingazi, Nemrut, Ambarli, Gemlik, Izmir, Marsaxlokk, and others (Figure~\ref{fig:traj_9160920}), with frequent crosses between North Africa, Turkey, and Malta. The $\text{SeqAcc}$ is 0\% for both validation and test sets, and performance degrades rapidly from 37\% at the first step to below 3\% at the third. Typical failures involve the model misplacing both the specific ports and their order, even if the predicted ports appear elsewhere in the vessel's history. Specifically, the model predicting 
Marsaxlokk~$\rightarrow$~Izmir~$\rightarrow$~Gemlik 
when the actual route is 
Bingazi~$\rightarrow$~Nemrut Bay~$\rightarrow$~Marsaxlokk. This finding suggests that the routing logic is influenced by ad-hoc business decisions rather than repeatable schedules, resulting in quasi-random and essentially unpredictable sequences, as evidenced by trajectory history.

The final category comprises vessels exhibiting a mixture of regularity and variability. These ships typically achieve moderate sequence accuracies, falling between the perfectly cyclic and the completely irregular cases. Their trajectories often follow a dominant loop or corridor but occasionally deviate due to side trips or port substitutions. IMO~8201624 (Figure~\ref{fig:8201624}) illustrates this pattern in the Indian Ocean. While the model correctly anticipates the overall operational corridor spanning Madagascar, Mauritius, and Reunion, it frequently misorders intermediate calls. For example, it might predict a direct return to Toamasina instead of a detour via Reunion. Consequently, the $\text{SeqAcc}$ remains in the mid-range, reaching approximately 12.5\% on the test set. Similarly, IMO~9231248 follows a long-distance trade between West Africa and North Europe. The model captures the global structure well and achieves a high accuracy of over 75\% at the first prediction step. However, $\text{SeqAcc}$ hovers around 37.5\% due to local variations, such as swapping adjacent ports or substituting functionally similar hubs within the same region. Taken together, these cases demonstrate that the model effectively exploits strong global structures but remains sensitive to local variations in port ordering. While the precise micro-sequence remains uncertain, the model reliably identifies the next region or main hub.

\section{Conclusion}\label{sec:conclusion}

This study addresses the critical challenge of multi-step port-of-call prediction in global liner shipping by proposing a generalized Connectivity-Constrained and Retrieval-Enhanced (CCRE) framework. To bridge the gap between short-horizon destination forecasting and long-horizon voyage visibility, we formulate the task as a joint sequence generation problem over a global maritime network graph. By utilizing high-resolution AIS data to overcome the limited accessibility and  unreliability of schedule information, the proposed framework provides a robust, data-driven solution for enhancing port tactical resource planning and carrier operational efficiency.

The CCRE network introduces several key methodological innovations to resolve the inherent complexities of maritime trajectory modeling. Inspired by the paradigm of RAG, we developed a retrieval-enhanced historical encoder that explicitly aggregates navigational precedents from a global database. By utilizing a dual-metric mechanism that combines Jaccard Similarity for global routing intent and Positional Match Rate for precise sequential alignment, this module effectively compensates for data sparsity in long-tail routes and resolves routing ambiguities. Furthermore, the architecture employs an adaptive feature-level ``middle fusion" strategy via a cross-attention module. This mechanism enables the model to dynamically shift its predictive focus across the forecasting horizon, relying on real-time kinematics for short-term accuracy while transitioning to historical contexts for long-term strategic stability.

To ensure superior sequence-level coherence and physical feasibility, we implement an autoregressive Transformer decoder augmented with Scheduled Sampling and Gumbel-Softmax relaxation. This joint sequence optimization facilitates end-to-end gradient propagation, which helps mitigate the error accumulation typical of traditional cascading models. Crucially, the decoding process incorporates topology reachability masks derived from the empirical shipping network to strictly enforce network connectivity and explicitly prevent the generation of operationally infeasible routes and explicitly prevent the generation of operationally infeasible routes.

Extensive empirical evaluations on a comprehensive global-scale liner shipping network demonstrate the framework’s superior effectiveness and generalization capability. The CCRE model achieves an accuracy of 72.3\% for the first destination prediction and an average accuracy of 61.4\% across a three-step prediction horizon. These results significantly outperform established baselines, including CatBoost and LSTM, by average margins of 12.6\% and 11.3\%, respectively. Feature ablation and qualitative case studies confirm the model’s scalability and its unique ability to capture both highly regular cyclic patterns and complex, partially regular trade lanes across diverse international corridors.

Despite these advancements, several promising avenues for future research remain. First, extending the framework into a continual learning paradigm would enhance its adaptability to changes caused by seasonal service reconfigurations or sudden network disruptions. Second, integrating external variables such as real-time port congestion indices, bunker prices, and severe weather indicators could further improve the interpretability of anomalous vessel behaviors. Finally, embedding these multi-step predictive sequences into downstream decision-support systems, such as uncertainty-aware estimated time of arrival forecasting, will be essential for realizing risk-aware, intelligent port operations.

\clearpage
\ACKNOWLEDGMENT{%
The acknowledgements section will be completed after the peer-review process.
}

%
\clearpage
\newpage
\section*{Declaration of generative AI and AI-assisted technologies in the writing process}
During the preparation of this work the authors used ChatGPT 5.2 in order to improve language and help write \LaTeX. After using this tool, the authors reviewed and edited the content as needed and take full responsibility for the content of the publication.
\newpage
\begin{APPENDIX}{}
\begin{algorithm}[H]
\small
\caption{Export sailing segments of AIS data}
\KwIn{Time-ordered AIS points $\{(\texttt{lat}_t,\texttt{lon}_t,\texttt{ts}_t)\}_{t=1}^{T}$ for one vessel; geofences $\mathcal{G}^{(\text{parking})}$, $\mathcal{G}^{(\text{pilot})}$, $\mathcal{G}^{(\text{berth})}$ (with \texttt{port\_name})}
\KwOut{AIS points in \textsc{sailing} only, each with \texttt{pre\_prtName}, \texttt{next\_prtName}}
\BlankLine
\textbf{Helpers:}\;
\Indp
$\mathrm{locate}(p)$:
\lIf{$p\in\mathcal{G}^{(\text{berth})}$}{\Return (\texttt{berth}, port\_name)}
\lElseIf{$p\in\mathcal{G}^{(\text{pilot})}$}{\Return (\texttt{pilot}, port\_name)}
\lElseIf{$p\in\mathcal{G}^{(\text{parking})}$}{\Return (\texttt{parking}, port\_name)}
\lElse{\Return (\texttt{none}, \texttt{None})}
\Indm
\textbf{Init:}\;
episodes $\leftarrow [\,]$; in\_episode $\leftarrow$ \texttt{False}; seen\_berth $\leftarrow$ \texttt{False}; start\_idx $\leftarrow -1$; cur\_port $\leftarrow$ \texttt{None}\;
\BlankLine
\For{$t=1$ \KwTo $T$}{ $(z_t, p_t) \leftarrow \mathrm{locate}((\texttt{lat}_t,\texttt{lon}_t))$;\ }
\For{$t=1$ \KwTo $T$}{
  \uIf{$\neg$ in\_episode \textbf{and} $z_t\in\{\texttt{parking},\texttt{pilot},\texttt{berth}\}$}{
    in\_episode $\leftarrow$ \texttt{True}; start\_idx $\leftarrow t$; cur\_port $\leftarrow p_t$; seen\_berth $\leftarrow$ ($z_t{=}\texttt{berth}$)\;
  }
  \uElseIf{in\_episode}{
    \If{$z_t=\texttt{berth}$}{ seen\_berth $\leftarrow$ \texttt{True}\; }
    \If{$z_t=\texttt{none}$}{
      \If{seen\_berth}{ append episodes $\leftarrow$ (start\_idx, $t{-}1$, cur\_port)\; }
      in\_episode $\leftarrow$ \texttt{False}; seen\_berth $\leftarrow$ \texttt{False}; cur\_port $\leftarrow$ \texttt{None}\;
    }
  }
}
\If{in\_episode \textbf{and} seen\_berth}{ append episodes $\leftarrow$ (start\_idx, $T$, cur\_port)\; }
\BlankLine
sailing\_intervals $\leftarrow$ complement of $\bigcup$episodes over $[1..T]$\;
\For{$j=1$ \KwTo $|$sailing\_intervals$|$}{
  let $(a_j,b_j)$ be the $j$-th interval\;
  pre $\leftarrow$ \texttt{None}; next $\leftarrow$ \texttt{None}\;
  \If{$j>1$}{ pre $\leftarrow$ episodes[$j{-}1$].port\_name\; }
  \If{$j\le |$episodes$|$}{ next $\leftarrow$ episodes[$j$].port\_name\; }
  \For{$t=a_j$ \KwTo $b_j$}{ AIS[$t$].\texttt{pre\_prtName} $\leftarrow$ pre;\quad AIS[$t$].\texttt{next\_prtName} $\leftarrow$ next\; }
}
\Return $\{\,\text{AIS}[t]\mid t\in \bigcup_j [a_j,b_j]\,\}$ in original time order\;
\end{algorithm}

\begin{algorithm}[H]
\small
\caption{Preprocessing producing $\big(\bm{X}^{(\mathrm{kin})}_s,\bm{X}^{(\mathrm{dyn})}_s,\bm{X}^{(\mathrm{stat})}_s,\bm{m}_s,\bm{h}_s,o_s,\bm{y}_s\big)$}
\KwIn{Per-vessel AIS CSVs; sampled IMO--static table; ports\_coordinates CSV}
\KwOut{Segments $s$ with variables in Table~\ref{tab:feature-schema}}
\ForEach{vessel file}{
  read CSV $\mathcal{D}$; cast \texttt{timestamp} to datetime; sort by time\;
  split consecutive blocks $B$ with non-null (\texttt{sailing\_start}, \texttt{sailing\_end}); discard $|B|<2$\;
  \ForEach{block $B$ in time order}{
    $L_s \leftarrow |B|$\;
    build $\bm{X}^{(\mathrm{kin})}_s\in\R^{L_s\times d_{\mathrm{traj}}}$ by stacking columns
    $[\texttt{lat},\texttt{lon},\texttt{speed},\texttt{course},\texttt{timestamp\_unix},\texttt{imo}]$ in time order\;
    set $\bm{m}_s\in\{0,1\}^{L_s}$ to $1$ on observed rows and $0$ on padding\;
    build $\bm{X}^{(\mathrm{dyn})}_s\in\R^{L_s\times d_{\mathrm{ves}}}$ from
    $(\texttt{speed},\texttt{draught},\texttt{heading},\texttt{course},\texttt{eta})$ aligned by time\;
    build $\bm{X}^{(\mathrm{stat})}_s\in\R^{d_{\mathrm{crr}}}$ from $(\texttt{length\_x},\texttt{width\_x},\texttt{TEU\_x},\texttt{crrId},\texttt{imo})$\;
    set $o_s \leftarrow$ first \texttt{pre\_prtName} (current departure port)\;
    record $t^{\mathrm{start}}_s \leftarrow$ first \texttt{timestamp} and store \texttt{imo/mmsi}\;
    append segment $s$ to vessel list $\mathcal{S}$\;
  }
  sort $\mathcal{S}$ by $t^{\mathrm{start}}_s$\;
  \For{$i=1$ \KwTo $|\mathcal{S}|$}{
    let $s_i$ be the $i$-th segment;\;
    set $\bm{h}_{s_i}$ and $\bm{y}_{s_i}$ with \texttt{None}-padding
  }
}
\nl build port set $\mathcal{P}$ from \texttt{ports\_coordinates} intersected with observed ports; construct mapping and adjacency $G$; precompute $G^h,\ h=1..M$\;
\nl for each $s$ and step $h$, form reachability mask $\bm{r}_{s,h}\in\{0,1\}^{|\mathcal{P}|}$ with $\bm{r}_{s,h}(j)=\mathbb{I}\!\big[(G^h)_{o_s,j}>0\big]$\;
\end{algorithm}

\begin{table}[htbp]
\centering
\caption{Baseline model architectures and hyperparameter settings.}
\label{tab:baselines}
\footnotesize 
\begin{tabularx}{\textwidth}{l X}
\toprule
\textbf{Model} & \textbf{Description \& Key Settings} \\
\midrule
\textbf{RF} & Bootstrap-aggregated decision-tree ensemble. \newline
\emph{Settings:} \texttt{n\_estimators}=200, \texttt{max\_depth}=10. \\
\addlinespace
\textbf{XGBoost} & Gradient-boosted decision trees (tree-wise additive). \newline
\emph{Settings:} \texttt{n\_est}=600, \texttt{lr}=0.1, \texttt{max\_depth}=8, \texttt{subsample}=0.9, \texttt{reg\_lambda}=1.0. \\
\addlinespace
\textbf{CatBoost} & Ordered boosting over symmetric trees. \newline
\emph{Settings:} \texttt{iters}=400, \texttt{lr}=0.08, \texttt{depth}=8, \texttt{l2\_leaf\_reg}=3.0. \\
\addlinespace
\textbf{LSTM/GRU} & Bidirectional recurrent encoders (2 layers, hidden 256). \newline
\emph{Settings:} Adam (\texttt{lr}=$10^{-3}$, \texttt{wd}=$10^{-5}$), 50 epochs, batch size 64. \\
\bottomrule
\end{tabularx}
\end{table}


\begin{table}[htbp]
\centering
\caption{Table of Notations}
\label{tab:notations}
\renewcommand{\arraystretch}{1.2} 
\begin{tabular}{@{}l l p{13cm}@{}}
\toprule
\textbf{Symbol} & \textbf{Type} & \textbf{Description} \\
\midrule

$\mathbf{A}$ & Matrix & Adjacency matrix of $\mathcal{G}$, where $[\mathbf{A}]_{ij} > 0$ indicates a link $i \to j$. \\
$C_{1:u}$ & Vector & Dynamic query context at prediction step $h$, concatenating observed history and generated predictions \\
$\mathcal{E}$ & Set & Set of edges representing historically observed direct sailings. \\
$\mathcal{G}$ & Graph & Maritime shipping network graph, defined as $\mathcal{G} = (\mathcal{P}, \mathcal{E})$. \\
$H$ & Scalar & Prediction horizon (length of the target sequence). \\
$\mathcal{H}$ & Set & Historical database established before training.\\
$L_s$ & Scalar & Length of the input historical trajectory sequence for sample $s$. \\
$\bm{m}_s$ & Vector & Binary validity mask, where $m_{s,t}=1$ indicates valid observed data. \\
$o_s$ & Scalar & Origin port (last observed valid port) for sample $s$, $o_s \in \mathcal{P}$. \\
$\mathcal{P}$ & Set & The set of valid port indices, defined as $\mathcal{P} = \{1, \dots, |\mathcal{P}|\}$. \\
$P^{(n)}_{1:u}$ & Vector & The prefix port sequence of the $n$-th historical scenario\\
$R^{(n)}_{1:H}$ & Vector & The subsequent ground-truth port sequence of the $n$-th historical scenario\\
$\mathcal{R}_h(o_s)$ & Set & Set of feasible candidate ports reachable from $o_s$ in exactly $h$ legs. \\
$\mathcal{S}_n$ & Tuple & The $n$-th historical navigation scenario retrieved from $\mathcal{H}$ \\
$\bm{U}_s$ & Matrix & Contextualized sequence representation output of Trajectory Encoder. \\
$u$ & Scalar & Effective context length at prediction step $h$, defined as $u = K + h - 1$ \\
$\mathbf{w}_{s,h}$ & Vector & Retrieval-prior weights over the top-\(N\) historical scenarios. \\
$\bm{X}^{(\mathrm{kin})}_s$ & Matrix & Dynamic trajectory kinematics feature matrix. \\
$\bm{X}^{(\mathrm{dyn})}_s$ & Matrix & Dynamic vessel operation feature matrix. \\
$\bm{X}^{(\mathrm{stat})}_s$ & Vector & Static carrier context vector. \\
$\mathcal{X}_s$ & Set & Comprehensive historical operational context for sample $s$. \\
$\mathbf{y}_s$ & Vector & Ground-truth future port sequence, $\mathbf{y}_s \in (\mathcal{P} \cup \{\omega\})^H$. \\
$y_{s,h}$ & Scalar & Port identifier at the $h$-th future step for sample $s$. \\
$y_{s, <h}$ & Vector & Sequence of ports predicted in steps preceding $h$, i.e., $(y_{s,1}, \dots, y_{s, h-1})$. \\
\(\mathbf{z}^{(\mathrm{traj})}_{s}\) & Vector & Trajectory representation of sample \(s\). \\
\(\mathbf{z}^{(\mathrm{hist})}_{s,h}\) & Matrix & Candidate-level historical representation at prediction step \(h\), with dimension \(N\times d_r\). \\
\(\mathbf{z}^{(\mathrm{fuse})}_{s,h}\) & Vector & Step-specific fused representation used for predicting \(y_{s,h}\). \\
$\theta$ & Param & Learnable parameters of the neural network model. \\
$\omega$ & Scalar & Sentinel token index for padding/unknown, defined as $|\mathcal{P}| + 1$. \\

\bottomrule
\end{tabular}
\end{table}
\end{APPENDIX}
%
%


\clearpage
\bibliographystyle{informs2014trsc} 
\bibliography{sample} 


\end{document}